Supplementary videos

# Origami-inspired soft twisting actuator

Diancheng LI[1,2], Dongliang FAN[1], Renjie ZHU[1], Qiaozhi LEI[3], Yuxuan LIAO[1], Xin YANG[1], Yang PAN[1], Zheng WANG[1,2,5], Yang WU[4], Sicong LIU[1], Hongqiang WANG[1,2,5*]

## Abstract

Soft actuators have shown great advantages in compliance and morphology matched for manipulation of delicate objects and inspection in a confined space. There is an unmet need for a soft actuator that can provide torsional motion to e.g. enlarge working space and increase degrees of freedom. Towards this goal, we present origami-inspired soft pneumatic actuators (OSPAs) made from silicone. The prototype can output a rotation of more than one revolution (up to 435°), more significant than its counterparts. Its rotation ratio (=rotation angle/ aspect ratio) is more than 136°, about twice the largest one in other literature. We describe the design and fabrication method, build the analytical model and simulation model, and analyze and optimize the parameters. Finally, we demonstrate the potentially extensive utility of the OSPAs through their integration into a gripper capable of simultaneously grasping and lifting fragile or flat objects, a versatile robot arm capable of picking and placing items at the right angle with the twisting actuators, and a soft snake robot capable of changing attitude and directions by torsion of the twisting actuators.

[1] Shenzhen Key Laboratory of Biomimetic Robotics and Intelligent Systems, Department of Mechanical and Energy Engineering, Southern University of Science and Technology, Shenzhen, 518055, China.

[2] Guangdong Provincial Key Laboratory of Human-Augmentation and Rehabilitation Robotics in Universities, Southern University of Science and Technology, Shenzhen, 518055, China.

[3] Education Center of Experiments and Innovations, Harbin Institute of Technology (Shenzhen), Shenzhen, 518055, China.

[4] School of Mechanical Engineering and Automation, Harbin Institute of Technology (Shenzhen), Shenzhen, 518055, China.

[5] Southern Marine Science and Engineering Guangdong Laboratory (Guangzhou), Guangzhou, 510642, China.

* Corresponding author. Email: wanghq6@sustech.edu.cn

## 1 Introduction

Soft robots, with high adaptability and deformation, are promising for safe human-robot interaction, dexterous manipulation, and inspection in a confined environment[1-4]. Although the bending, expanding, and shrinking movements of soft robots have been widely studied, little research exists on twisting (torsional) soft actuators[5]. Rotation along the actuator axis is usually one of the primary degrees of freedom (DOFs) in robots and their natural counterparts. For example, the rotation of the necks enables us to acquire information in a larger scope. Most previous rigid robots possess axial rotation motions for dexterous manipulation[6]. Similarly, various scenarios also require twisting motions in soft robots. For instance, the soft rotational structure was also necessary to simulate the biomorphology of the left ventricle of the heart[7]. Lacking torsion motion, for example, a soft robotic arm can hardly adjust to the suitable angle to complete the grasping task[8], and a soft snake robot is difficult to change the view attitude or the operating angle.

It is only recently that researchers have developed soft twisting actuators. For example, Sanan et al.[9] developed three types of soft actuators, based on a double or triple helix structure made of non-extendable fabric material, a rotary peano actuator, and multiple McKibben actuators, respectively. The maximum rotation angle among these reached 45°. Kurumaya et al.[8] and Connolly et al.[10] reported a class of inflatable twisting actuators using non-stretchable fibers as the confining layer, and by optimizing the fiber angle, these actuators achieved a rotation angle up to 195°. Yang[11] built a four-chamber design using a combination of two different shaped chambers, which could perform a twisting angle of 30°. Jiao et al.[12] designed a single-chamber actuator with six faces to achieve a maximum of 80° of torsion through the collapse in the central region during deflating. However, very few soft-twist actuators that can reach one turn of twist (i.e., 360°) have been developed. Theoretically, some existing soft-twi st actuators can increase the rotation angle by simply stacking more modules. Still, this solution unfavorably increases the size and the response time.

Here we present a novel soft torsional actuator by introducing origami-structured silicone shell. Although previously soft origami robots have been studied extensively[13,14], there are few designs for rotation of soft actuators as far as we know. Owing to its large transformation ratio, the origami structure potentially allows a larger and programmable rotation angle in the twist actuator[15-17]. In this work, we integrate Kresling pattern[18] into the design of the actuator. With pressured air and vacuum, the actuator unfolds and folds, respectively, resulting in the twist of the actuators. There have been some attempts to design robots using Kresling origami[19-22]. However, they are not specifically designed for rotation actuators or soft actuators[23-29].

The contributions of this work are as follows. First, this paper built the analytical models for the OSPAs and parametrically analyzed the design parameters to optimize the maximal rotation angle. Also, we established the kinetic model in terms of air pressure and torsion angle. We fabricated four OSPAs of different types using silicone instead of paper based on inner dip coating method. The silicone-based OSPAs are more reliable and robust, and they survived in our stretching and compressing test (up to 150% stretching ratio), load-bearing test (2 kg load), and the impact test (hammer knocking), while the paper-made counterparts buckled or collapsed during these tests. The maximum rotation angle of the OSPA prototypes made in this work approached 435°, larger than most of the previous soft twist actuators as far as we know[8-12,23-29]. The corresponding actuator's rotation ratio (the rotation angle to the aspect ratio) is 136°, about two times higher than the maximum value in the literature[28]. Furthermore, by combining these actuators, we demonstrated the vast applications of the OSPAs in three different robots. The first was a soft gripper with in-plane rotation motion, capable of picking fragile items (e.g., a cherry tomato) and thin films (e.g., a piece of paper) by gripping and lifting simultaneously with the assistance of soft rolling. The second prototype was a versatile robotic arm. Connected with a camera, it changed the camera's attitude and perspective to monitor a robotic fish that was randomly swimming in the water. With a suction cup at the free end, this soft arm successfully picked and placed items into the corresponding slots, based on the robot's shrinking, expanding, and twisting motions. The OSPA snake robot finally demonstrated compliant movements in a pipeline and the spinning function of the head owing to the twisting of the OSPAs.

The structure of this paper is as follows. The next section introduces the design principle, models, and parametric analysis of the actuators. Section 3 displays and discusses the experimental results in this work.

# 2 Methods

## 2.1 Design

In many cases, the limited reachable rotation angle restricts the design of soft robots[5,10,23,30-32]. In this work, inspired by the structure of the Kresling pattern, origami spring, and twisted origami towers[19-21,33,34]. We designed origami-inspired soft pneumatic actuators (OSPAs) that are capable of twisting for a large angle. Each OSPA combines multiple OSPA modules in a series, as shown in Fig. 1. An OSPA module is in Kresling pattern, and it possesses two identical polygons in parallel as the upper and bottom sides, and the surface intersected with the creases (the lines connecting the vertices of the upper and bottom polygons) as the walls.

Being vacuumed or inflated, an OSPA module deforms, and the free end of the module moves linearly and rotates simultaneously with respect to the fixed end. According to the angle of the creases, the rotation direction of the OSPA module can be clockwise (CW) or counterclockwise (CCW). By combining the modules together, we can have distinct output motions on the free ends, as shown in Fig. 1. For example, two modules of the same rotation direction connected generate a rotation and linear stroke on the free end, while two modules of the opposite directions create only linear output motion on the free end (no rotation). Similarly, modules of more than three connected together generate three types of output motion primarily, as shown in Fig. 1. If all the modules are in the same rotation direction (Type I), the free end rotates and shrinks while being vacuumed. Comparably, if half of the actuator is composed of modules of one rotation direction, the other half is modules of the opposite rotation direction, the free end only contracts (Type II). Finally, if each module has the opposite rotation direction with respect to the mirrored module (Type III), the actuator also generates only linear motion at the free end. In Type III, parts of the connection edges between the modules rotate, and parts not, while Type II has all the connection edges rotate, among which the middle one generates the largest rotation angle. By integrating these OSPAs of different types, robots of distinct functions can be conceived for extensive applications, such as a gripper, a robotic arm, and a snake robot, as demonstrated in the results section. It is noted that rotation in some of the actuators (Type II and III) might not be necessary for particular scenarios. They can be replaced by traditional soft elongating actuators too, to avoid the potential energy loss in the rotation part.

Moreover, different from most of the previous origami-inspired robots made from paper, the OSPAs in this work were made from elastomer to enhance the compliance and robustness of the robot. The detailed fabrication process is explained in the Fabrication section. Its comparison with the paper-made structure can be found in the results section.

## 2.2 Analysis

### a) Analytical model

We assume that *a, b, c* are the lengths of the creases of the modules, as shown in Fig. 2(A). $\delta$ is the folding angle between the hypotenuse $b$ and the horizontal plane. Point $N$ is the projection of a vertex of the upper hexagon on the horizontal plane. Point $O$ is the center of the bottom polygon. The angle between the $OM$ and $ON$ lines is $\theta_u$, which represents the rotation angle of the upper hexagon along the vertical axis during the module folding process. For the regular hexagons with the circumscribed circle of radius $r$, there is,

$$OP = OM = r = a\,. \tag{1}$$

We know the distance between the upper and bottom hexagons is,

$$h = b\sin\delta\,. \tag{2}$$

Then, when the structure is folded, the relative rotation angle between $OM$ and $ON$ can be achieved by,

$$\theta_u = 2\arcsin\frac{\sqrt{b^2-h^2}}{2a} = 2\arcsin\frac{b\cos\delta}{2a}\,. \tag{3}$$

Hence, the relative rotation angle $\theta_u$ is primarily decided by the angle $\delta$ and the ratio $b/a$. As shown in Fig. 2(B), $b/a$ increases the relative rotation angle, but the angle $\delta$ decreases it.

Moreover, the maximum rotation angle during the folding of the module is expressed by,

$$\theta_f = 2\arcsin\frac{b}{2a} - 2\arcsin\frac{b\cos\delta}{2a}\,. \tag{4}$$

In this case, both $b/a$ and the angle $\delta$ rise the folding rotation angle $\theta_f$, as shown in Fig. 2(C). The total maximum rotation angle of the module can be, if e.g., the module unfolds first and then folds,

$$\theta_{max} = \theta_u + \theta_f = 2\arcsin\frac{b}{2a}\,. \tag{5}$$

This equation indicates that the maximum rotation angle of the module has no relationship with the initial angle of the ridges ($\delta$), but it is only decided by $b/a$ of the module. The above analyses also reveal that the number of sides of the regular polygon (only if it is larger than 3) has no effect on the rotation angle. More sides enlarge the upper and bottom polygon area, and therefore are more likely to increase the force, but they unfavorably raise the fabrication difficulties simultaneously. In this work, we chose hexagons for the upper and bottom polygons of the modules as a tradeoff. This model is built for a module, based on which the kinematics model for an actuator with a series of modules can be developed (see Supplementary Data for details on kinematics of an OSPA actuator composed of a series of modules).

A kinetic model is established to describe the relationship between the rotation angle $\theta_u$ and the internal pressure $p$ as follows. If there is no load, according to the principle of virtual work, the sum of the work related to internal pressure $p$: ($\partial W_P$) and the stored energy of the system ($\partial W_S$) is zero for any infinitesimal virtual displacements[23],

$$\partial W_P + \partial W_S = 0. \tag{6}$$

The work done by the external force can be expressed as the work of the air in the inner chamber,

$$W_p = pV_C, \tag{7}$$

where $V_C$ denotes the internal volume of an OSPA module. Based on the finite volume method[35], the volume of a polyhedral OSPA module can be calculated from the sum volume of a number of polygonal pyramids, where geometric center $G$ is its vertex. Thus the internal volume of an OSPA module can be given as,

$$V_C = \sum_{n_{pyramid}} V_{pyramid}. \tag{8}$$

The stored elastic energy in an OSPA module can be expressed by[36-39],

$$\begin{aligned} W_{S1} = {} & \frac{k_{c1}}{2}\sum_{i=1}^{n}\left[2a\left(\theta_{QR,i}-\theta_{QR,0}\right)^2 + 2b\left(\theta_{QM,i}-\theta_{QM,0}\right)^2\right] \\ & + \frac{k_{c2}}{2}\sum_{i=1}^{n}\left[c\left(\theta_{RM,i}-\theta_{RM,0}\right)^2\right], \end{aligned} \tag{9}$$

where $k_{c1}$ and $k_{c2}$ are the torsional elastic constants corresponding to different thicknesses (see Supplementary Data for details on flexural rigidity and elastic constants). $\theta_{QR}$, $\theta_{QM}$, $\theta_{RM}$ are the rotation angles of the folds QR, QM, RM respectively and the rotation angle of the folds can be found in Supplementary Data Fig. S2 and Fig. S3 (see also Supplementary Data for volume of the OSPA module and the rotation angle of the folds). According to Equation (6), we can achieve the relationship between the rotation angle and the pressure.

Using the principle of virtual work, we can also calculate the external torque $\tau$ as a function of relative rotation angle $\theta_u$ [9],

$$\partial W_\tau = \partial W_P + \partial W_S, \tag{10}$$

where $\partial W_\tau$ is the virtual work associated with the external torque $\tau$. Thus, we have,

$$\tau(\theta_u) = p\frac{dV_C}{d\theta_u} + \frac{dW_S}{d\theta_u}. \tag{11}$$

## b) Simulation model based on FEA

To verify the above kinematics model and analyze the influence of materials on the actuator's output force, a numerical study is performed using a finite element analysis (FEA) model (Abaqus, version 2020, SIMULIA, Dassault Systèmes).

The OSPAs were modeled as an incompressible isotropic hyperelastic material. The properties of the materials in the simulation of the OSPA module correspond to three different materials, including E615 silicone (Hongyejie Inc), Ecoflex 00-30, and the mixture of Dragonskin 30 and Ecoflex 00-30 with a mass ratio of 1:1. Their specifications are listed in Table S1 achieved from tests. The stress-strain curves of these materials are highly nonlinear and, therefore, cannot rely solely on the elastic modulus of the materials to characterize their behaviors when they are in large deformation. Referring to the ASTM D412 rubber tensile test, we conducted uniaxial tensile tests on each material. A more generalized function of Yeoh model is[40],

$$W = \sum_{i=1}^{N} c_{i0}(I_1 - 3)^i, \tag{12}$$

where $C_{i0}$ are material constants, and $I_1$ is the first strain invariants (axial, circumferential and radial) of the Cauchy-Green deformation tensor. The mixed component of Dragonskin 00-30 and Ecoflex 00-30 with a 1:1 weight ratio gives a favorable linear stress-strain curve, which will facilitate the modeling and control. The relationship between $I_1$ and principal stretch ratios $\lambda_1$, $\lambda_2$, and $\lambda_3$ can be expressed as,

$$I_1 = \lambda_1^2 + \lambda_2^2 + \lambda_3^2. \tag{13}$$

The OSPAs for simulation consisted of walls and top/bottom plates. The load was applied uniformly on the inner wall of the model which was meshed with tetrahedral elements. This equation was fitted using the obtained stress-strain curves (see Fig. S4 and Fig. S5). Then we input the fitted three parameters into Abaqus/explicit, divided the mesh and ran the job (Investigation of morphing by FEA can be found in Fig. S6 and Fig. S7).

We used FEA simulations to compare the variation of the rotation angle of OSPAs made of three different materials with the pressure (See Fig. S8). At first, the angle rises slowly with the pressure, and then it grows drastically, and after a threshold, it rises gently again. This tendency is true for different materials.

In Fig. 2(E), we compared the rotation angle from the FEA simulation model, the analytical model, and experiment as a function of $b/a$. The calculation results are consistent with the experimental data. Higher discrepancy existing at high $b/a$ perhaps because the enlarging of the panels becomes more dominant. Based on the these models, we also analyzed the influence of the pressure on the rotation angle.

As shown in Fig. 2(F), we compared the results of the FEA simulation model, the analytical model and the experiment as a function of the pressure. The simulation data share a similar tendency with the experimental results, and the magnitude deference perhaps comes from the inaccuracy on material parameters and the collision calculation of the inner walls in the simulation model. The analytical results agree well with the experimental data for the pressure ranging from -0.6 kPa to 0 kPa (corresponding to most of the generated rotation angles, from -68° to 0°), while they largely differ from the experimental results at other pressures perhaps since the analytical model lacks considering the loss of energy from the collision of the inner walls while being deflated and stretching on the panels while being inflated.

### 2.3 Fabrication

The manufacturing of the OSPAs utilized spin casting and slush casting. First, we 3D printed the molds, as shown in Fig. 3. Then we filled it with the uncured silicone and rotated the mold at the speed of 10 r/min to coat the silicone homogeneously on the inner mold wall (see Fig. 3(B)). After the silicone viscosity increased and the thickness of the silicone became stable, we cured the silicone by heating it in an oven at 45°C for 30 mins. Finally, the elastomer shell was peeled off from the mold. A more detailed manufacturing process is available in Table S2.

The soft elastomer endows the OSPAs distinguished strain ratio compared with the paper-made similar structure, as discussed in the experimental section in detail. However, the elastomer shells suffer buckling while the actuators are vacuumed since the creases should bear heavy axial loading, which is challenging when the elastomer wall is thin. To avoid failure, we intentionally strengthened the thickness of the creases and the edges just like origami tower. As shown in Fig. 3(D) and (E), in the reconstructed structure acquired by a 3D scanner (Handyscan portable 3D laser scanner, visualization by VxScan), the thickness around the crease and edges (e1 and e2 region in Fig. 3(E)) is approximately three times larger than those on other regions.

In this fabrication process, we can change the shape of the modules by changing the unfolding angle of the mold. By increasing or decreasing the duration of spin coatings, we can obtain different panel thicknesses, corresponding to different rigidity. According to the analyses and application requirements, this work made four OPSAs of different parameters, as shown in Table 1. By combining these actuators, we created various robots as described in the Demontration section.

# 3 Results

## 3.1 Characterization

### a) Rotation angle

To measure the rotation angle of the actuator, we fixed the top of the actuator (Actuator IB, $\delta = 53^{\circ}$, $b/a = 2$) and mounted a red pointer at the bottom as shown in Fig. 4(A). We deflated and then inflated the actuators to observe the rotation angle. The rotation angle during this process was consistent with the prediction from the FEA model and the theoretical model, as shown in Fig. 2(E). The slight discrepancy among them perhaps resulted from the larger thickness on the mountain crease regions (e1) as shown in Fig. 3(E). All these results show little hysteresis when the actuators deflate and inflate, which is beneficial for precise control.

With the same setup, we also measured the influence of the pressure. As shown in Fig. 4(B), although overall the rotation angle rises nonlinearly with respect to the pressure, some sections could be regarded as a linear region for control. For example, for Actuator IB, its linearity was 3.828% ranging from -250° to 0°. The maximum rotation angle of the actuator was 435° for Actuator IB (210° for Actuator IA, see Fig. 4(C), Fig. S9 and Video S1). This value is much higher than most other soft twist actuators, as shown in Fig. 4(D). Since the aspect ratio directly influences the rotation angle, to compare the soft twisting actuators fairly, we normalized the rotation angle by the aspect ratio of the actuators. By comparing the rotation ratios ($E_r$) as shown in Fig. 4(E), we found our actuator is the champion (136°) among all the soft torsional actuators as far as we know, approximately two times higher than the secondary one[28] (see also Table 2).

### b) Torque and torsional rigidity

We tested the torque $T$ of the OSPAs with the results shown in Fig. 4(F). We hanged Actuator IB on a frame, and the bottom free end pushed a load cell (FUTEK LSB200) by a stick while being deflated. The data of the load cell was obtained by LabVIEW through the National Instruments acquisition card. The internal air pressure was kept at -5 kPa by the controller with feedback from the pressure sensor. The same actuator at different operating lengths was tested. As shown in Fig. 4(G), when the operating length of the actuator became larger, the torque increased first and then dropped. The maximum torque (24 N·mm) occurred when the operating length was 108.4 mm (experimental setup for torque testing can be found in Fig. S10).

We calculated the torsional rigidity by,

$$K = \frac{Tl}{\varphi' - \varphi_0}, \tag{14}$$

where $l$ is the operating length of the actuators, $\varphi_0$ is the original end rotation angle, and $\varphi'$ is the consequent angle after the operation. As shown in Fig. 4(G), while being deflated, the torsional rigidity approaches 212.7 N·mm$^2$/° at maximum at an operating length of 108.4 mm. Based on Equation (11), we calculated the torque and rigidity of the analytical model and the trend is consistent with the measured torque and rigidity, respectively, as shown in Fig. 4(F) and (G). The errors between the experiments and analytical models may arise from the stretching of the silicone panels which absorb some of the energy at larger negative pressures, and those become more significant at larger operating lengths.

**c) Adaptability and robustness**

To test the adaptability, we compared the OSPAs made of silicone and those with the same geometry but made of paper, which are popular in previous literature[28,29]. As shown in Fig. 5(A), Actuator IA was stretched and compressed by a testing machine (MTS Model 42). There was no noticeable damage after it was stretched for more than 160%. As a comparison, the paper-made counterpart (paper: 180 g/m$^2$) collapsed on some of the valleys when the stretch ratio was approximately 122%, and it had fractures on the paper when the stretch ratio approached 140%. Therefore, the great adaptability of the OSPAs results from both the origami structure and the elastomer material. It is more suitable for soft robotic applications.

Then we also tested the robustness of the OSPAs by a heavy load (2 kg), as shown in Fig. 5(B). The OSPAs made from elastomer recovered soon, but the paper structure failed to return to its original shape. Again, after being sharply knocked by a hammer, the OSPA survived, but the paper-made structure did not (see Supplementary Video S2).

**d) Power and Efficiency**

The experimental setup for the efficiency test is shown in Fig. 5(D). The OSPA (Actuator IB) was inflated by a large syringe which was driven by a linear motion stage (Suhen Inc). The actuator was placed horizontally. One of its ends was only linearly movable, and the other end could only rotate. A weight was attached to the rotary end by a nylon wire.

The maximum output power generated by the torsional motion is about 7.67 mW, and the maximum power density is about 0.51 W/kg. The energy efficiency can be calculated by the equation[41,42],

$$\xi_t = \frac{T\theta}{PV_C}, \tag{15}$$

where $\theta$ is the torsion angle of change during the tests and $P$ is the effort variable of the pressure. The average efficiency for the weight of 5 g, 10 g, and 15 g was 7.50%, 10.32%, and 2.82%, respectively, which is comparable to the values of previous soft actuators[41,42].

**e) Reliability**

To test the reliability, we also utilized the setup shown in Fig. 5(D). The syringe infused into and withdrew a certain volume of air from the actuator repeatedly, driven by the motor. The sensor measured the pressure in the actuator, and this process was also filmed by a camera. As shown in Fig. 5(E), before and after 1000 cycles of shrinking and elongation, the pressures in the actuator kept almost the same, indicating that no fracture in the actuator's structure results from the long-term operation. The actuator for the test had been used for tens of hours of experiments on the robot arm and stored for over a year prior to the reliability test. Moreover, we did not find any collapse on the actuator too. Therefore, these actuators possess excellent reliability.

## 3.2 Demonstration

To demonstrate the extensive applications of the OSPAs, we designed three distinct robots integrating the OSPAs of different types.

**a) Gripper**

Gripping is extensively desired in automated assembly operations. Inspired by the dexterous human hands, researchers have developed a series of robotic hands[43-49]. A dexterous gripper with actively driven rollers located at the fingertips and generating lateral motions on the grasped objects inside the gripper is a promising approach[50]. Nevertheless, few soft hands have such a helpful skill.

In this work, using Actuator II, we built a rotatable soft gripper. This prototype had a compliant body like other soft grippers, while it could generate a rotation motion different from most previous soft grippers. As shown in Fig. 6(A), the gripper was composed of two OSPAs (Actuator II), a slider, and a supporting frame. The actuator ends could slide straight along the linear bearings. Noted that only about three-quarters of each actuator end was bonded to the slider to allow bending of the actuator, as shown in Fig. 6(B).

While inflated, as shown in Fig. 6(B), the actuators elongated, and its center section rotated and bent outward due to the limitation in the length direction (the gripper opened). After being deflated, the two actuators shrunk and rotated in the opposite direction, and the gripper closed. The gripper could clamp objects at this moment. This capability is advantageous for picking up a small and fragile object or a flat and flexible object. As shown in Fig. 6(C), the gripper gently grasped a cherry tomato (approximately 30 mm in diameter) using the rotation motion of the OSPAs. Based on our test, this gripper could grip and lift objects simultaneously. The object size was up to 50 mm in diameter and weight was over 500 g. This gripper also picked up a piece of paper and plastic film by friction and in-plane motion resulting from the rotation (see Supplementary Video S3), while most previous soft grippers were inadequate[50-53].

Also in Fig. 7(B), we demonstrated the lifting capability of the gripper, which can transport the strawberry from the lower surface to the upper surface during the gripping process. This provides a more stable gripping mode, making the strawberry less likely to fall from gripping and avoiding damage to the strawberry from the stress of prolonged gripping.

**b) Versatile deployable robot arm**

Previously, most soft robotic arm can only bend and shrink, and only few of them can rotate along the body axis, which is a critical function for various scenarios[8,54]. With the torsional soft actuators that can generate a rotary angle more than anyone in previous literature (see Table 2), here we built a soft robotic arm with three types of the OSPAs (Actuator III, IA, and IB with internal support skeleton), which was staged on a linear sliding table. With the compensation of Actuator III in the elongation, the end effector could twist (no expanding/shrinking), or twist and expand/shrink simultaneously (detailed information about the internal support skeleton can be found in Supplementary Data and Fig. S12).

We demonstrated this soft robotic arm on a monitor, as shown in Fig. 8(A). We installed a waterproof camera at the end of the robotic arm. The camera on the twist actuator (Actuator IB), with a maximum rotation angle (435°) of more than 360°, observed the movement of the robot fish in all directions by rotating and following the fish's position (see Supplementary Video S4).

Mounted a suction cup on the end effector, the OSPA robotic arm was able to pick and place items. By selectively controlling the pressure in the two OSPAs, the robotic arm could rotate or lift objects. Here the soft robotic arm successfully picked and placed an egg in the right direction, stacked blocks at the aimed position, and picked and inserted items of different shapes at the right slots, respectively, as shown in Fig. 8(B - D). These demonstrations used Actuator IA, which has a maximum rotation angle of about 120°. These demonstrations also show that the soft robotic arm tolerated the little position errors with its compliance, which is superior to the rigid robotic arms.

### c) Soft snake robot

Various soft snake robots and pipe crawling robots have been developed in the last decades for applications, including disaster rescue[55-60]. However, most of them lack the degree of axial rotation, which is critical for the versatility of robots.

As shown in Fig. 9(A and B), the OSPA snake robot developed in this work combined four OSPAs (Actuator IA and mirrored Actuator IA) in series, an analog camera (Chuangxinda Inc.), and a microphone. The first two actuators and the last actuator had a clockwise (CW) structure, and the third actuator had a counterclockwise (CCW) rotating structure. When the CCW actuator and one of the CW actuators operated simultaneously, the robot expanded or shrank but not rotated. If only one of the CCW/CW actuators worked, the robot head spun (See Supplementary Video S5). These actuators were connected to the pneumatic sources through solenoid valves (CONJOIN CJV23), which were controlled by a microcontroller board, as shown in Fig. S13 and Fig. 9(C).

To simulate the inspection task in a pipe network using the OSPA snake robot, we set up a platform with two exits (A and B) using PVC pipes and 3D printed frames, as shown in Fig. 9(D). By intentionally controlling the pressure of each OSPA, the soft snake robot shrank and expanded in sequence to pass through the complex pipe to the exit. The snake robot moved forward by being inflated and deflated repeatedly. Twisting helped the robot to turn the head to the right direction in the pipe, which was critical at the fork. As shown in Fig. 9(C and E), with the rotary actuator, the snake robot head rotated from one exit to the other, which means the soft robot could actively select the path while facing a divergence.

## 4 Summary

This work presents an origami-inspired soft twisting actuator made from silicone. The analytical model is built and validated by the FEA model and experimental results. These models indicate that the rotation angle of the OPSAs increases with the folding angle $\delta$ and the aspect ratio $b/a$ and highly nonlinearly grows with the air pressure. Based on the analysis, we fabricated various prototypes, which demonstrated a super large rotation angle of 435° (the corresponding angle ratio is 136.4°, superior to any previous counterpart), high stretchability (up to 160%), robustness (surviving from heavy load and harmer knocking), and great reliability (no performance decay after 1000-cycles operation). Moreover, the prototypes generated a maximum torque of 24 N·mm with the torsional rigidity of 212.7 N·mm$^2$/° with an efficiency up to 10.32%. With the actuators, three types of robots were implemented, including a soft gripper capable of grasping and lifting with the rotation of the soft twisting actuators, a robotic arm capable of picking and placing items to the right place and the right angle with the twisting motion, and a snake robot that was able to rotate the moving direction with the twisting actuators.

The actuators in this work mainly generate rotation and elongation, but they can also generate bending if the length of one side is limited. Their design, modeling, and applications will be studied in detail in the future. Other features (e.g., bi-stability) and different driving methods (e.g., dielectric elastomer actuation) will also be considered in future work.

## Acknowledgments

This work was supported in part by the National Natural Science Foundation of China under Grant 51905256, in part by the Natural Science Foundation of Guangdong Province of China under Grant 2020A1515010955, in part by the Science, Technology and Innovation Commission of Shenzhen Municipality under Grant ZDSYS20200811143601004, in part by the Natural Science Foundation of Liaoning Province of China (State Key Laboratory of Robotics joint funding, under Grant 2021-KF-22-11), and in part by Southern Marine Science and Engineering Guangdong Laboratory (Guangzhou) under Grant K19313901.

## Author Disclosure Statement

No competing financial interests exist.

# Tables

**Table 1.** Specifications of OPSA prototypes' mold

| **Actuator** | **Type** | ***b/a*** | $\delta_0$ | ***a*/mm** | **c/mm** | $h_0$**/mm** | ***n* (Number of modules)** |
|---|---|---|---|---|---|---|---|
| IA | I | 1 | $45^{\circ}$ | 18 | 27.6 | 12.7 | 8 |
| II | II | 1 | $45^{\circ}$ | 18 | 27.6 | 12.7 | 8 |
| III | III | 1 | $45^{\circ}$ | 18 | 27.6 | 12.7 | 8 |
| IB | I | 2 | $53^{\circ}$ | 20 | 44.7 | 31.9 | 4 |

**Table 2.** Comparison between different fluid-drive soft twisting actuators with a single input

| Actuator | Rotation Angle (°) | Aspect Ratio (Rest Status) | Internal Pressure Change (kPa) | Maximum Extension/ Shrinkage | $E_r$ (°)* |
|---|---|---|---|---|---|
| Kurumaya et al. | 77 | ~2.55 | 172 (0 to 172) | - | 30.2 |
| Connolly et al. | ~195 | ~8.6 | ~62 (0 to 62) | ~3% | 22.7 |
| Yang et al. | ~30 | ~1 | 5 (-5 to 0) | - | 30.0 |
| Jiao et al. | ~80 | ~1.33 | ~70 (-70 to 0) | ~-40% | 60.0 |
| Yan et al. | ~100 | 2.78 | ~100 (0 to 100) | ~0% | 35.9 |
| Morin et al. | ~20 | ~3 | ~40 (0 to 40) | ~7.5% | 6.7 |
| Gorissen et al. | ~70 | 1.57 | ~178 (0 to 178) | - | 44.6 |
| Belding et al. | ~175 | ~10.81 | ~360 (0 to 360) | - | 16.2 |
| Lazarus et al. | ~65 | ~2.9 | ~38 (-38 to 0) | - | 22.4 |
| Martinez et al. | ~415 | ~5.76 | ~12 (0 to 12) | ~17% | 72.0 |
| Li et al. | ~90 | ~6 | ~60 (-60 to 0) | - | 15.0 |
| Sanan et al. | ~45 | > 2 | - | - | < 30 |
| This work | 435 | ~3.19 | ~12 (-10 to 2) | ~-78% to 9% | 136.4 |

*Rotation ratio $E_r = \frac{Rotation\ \ Angle}{Aspect\ \ Ratio}$.

## Figures

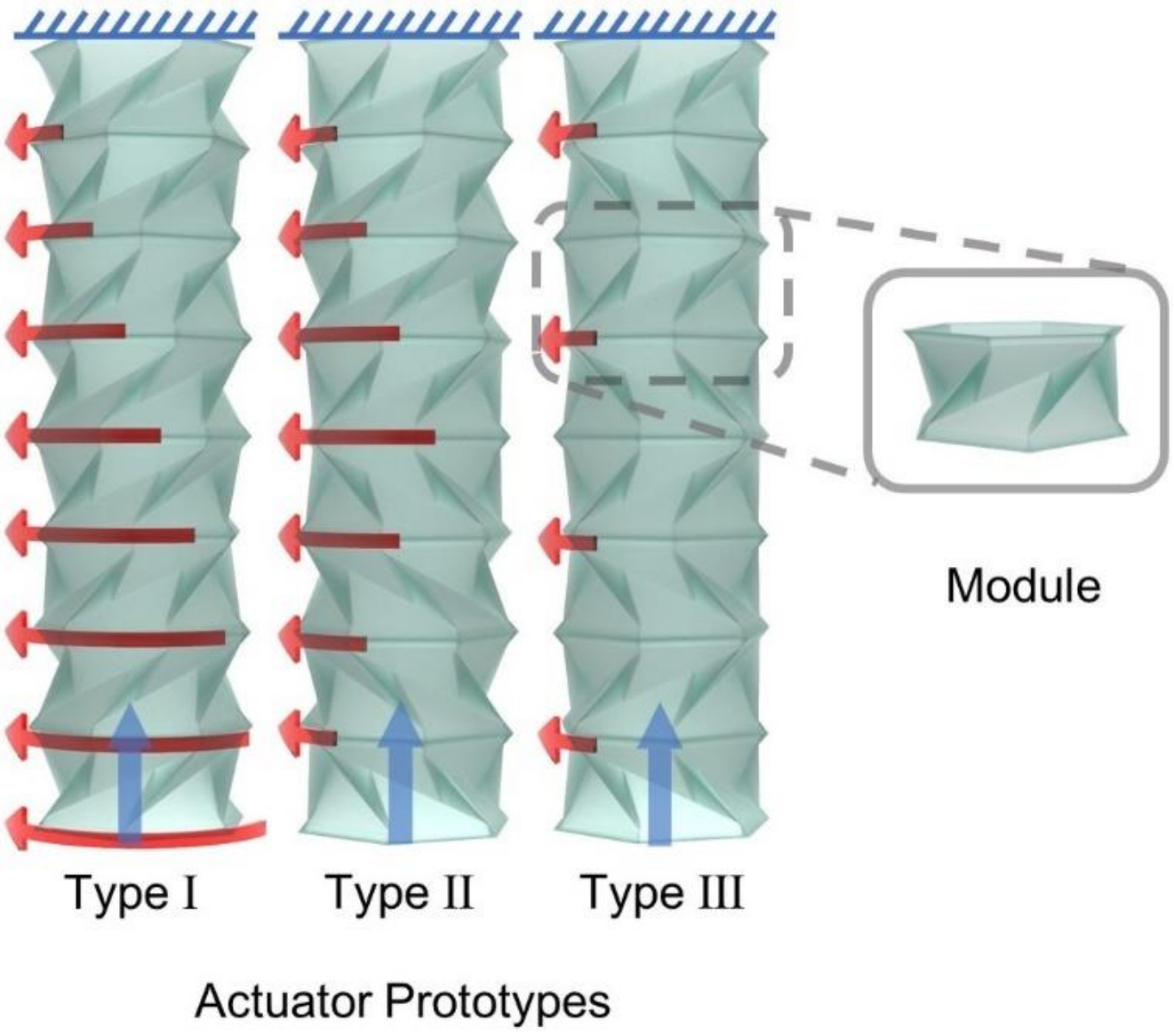

**FIG. 1.** The concept and structures of the OSPAs.

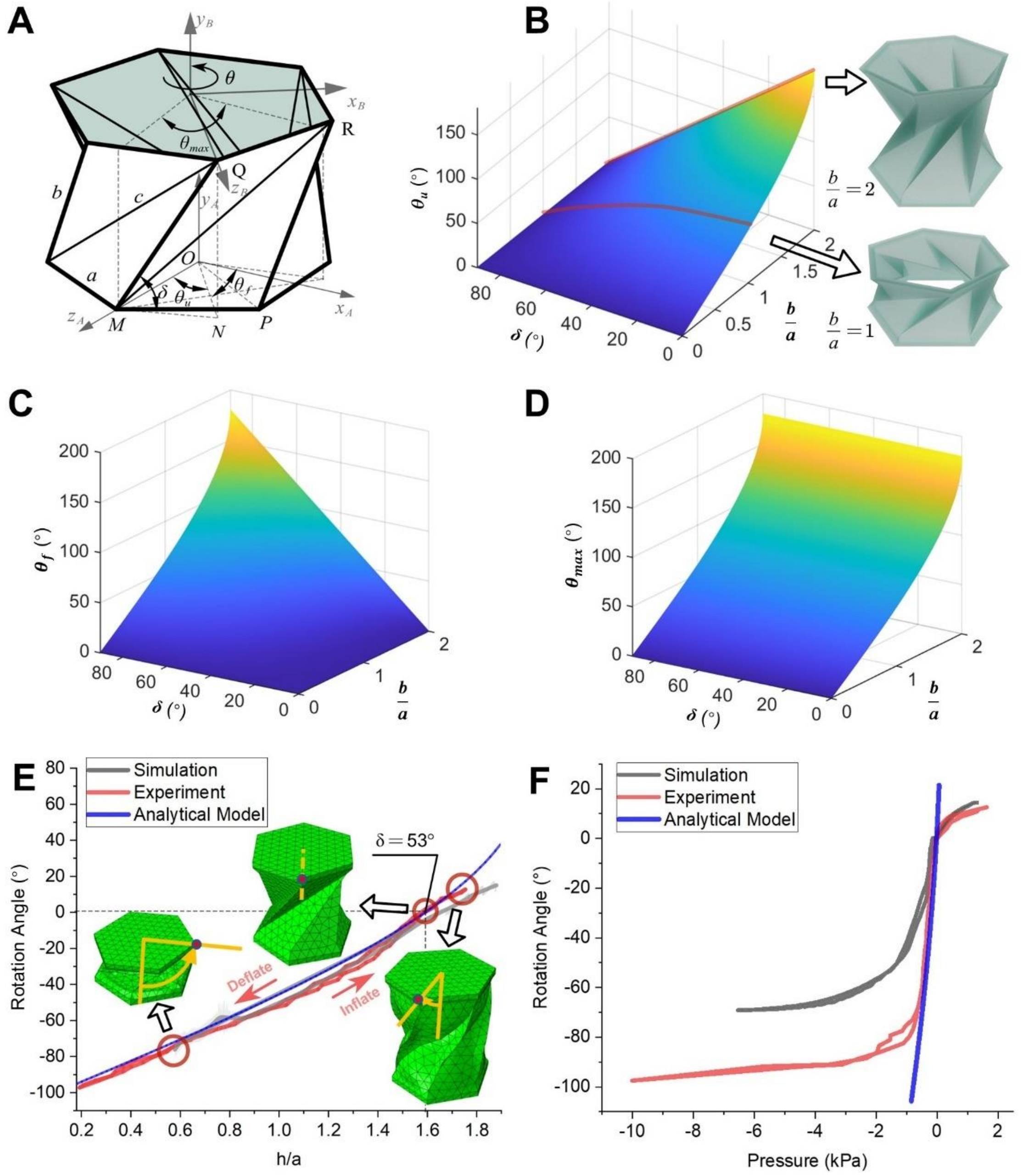

**FIG. 2.** Analysis on the OSPA modules. (A) The geometric parameters of an OSPA module. (B) The influence of $\delta$ and *b/a* on the rotation angle $\theta_u$ during unfolding. (C) The maximum rotation angle $\theta_f$ of the OSPA module during deflation. (D) Total maximum rotation angle $\theta_{max}$ of OSPA module. (E) Comparison of the rotation angle resulted from analytical model, simulation, and experiment with respect to *h*/*a* (Type IB module). (F) Comparison of the rotation angle resulted from analytical model, simulation, and experiment with respect to different pressure (Type IB module).

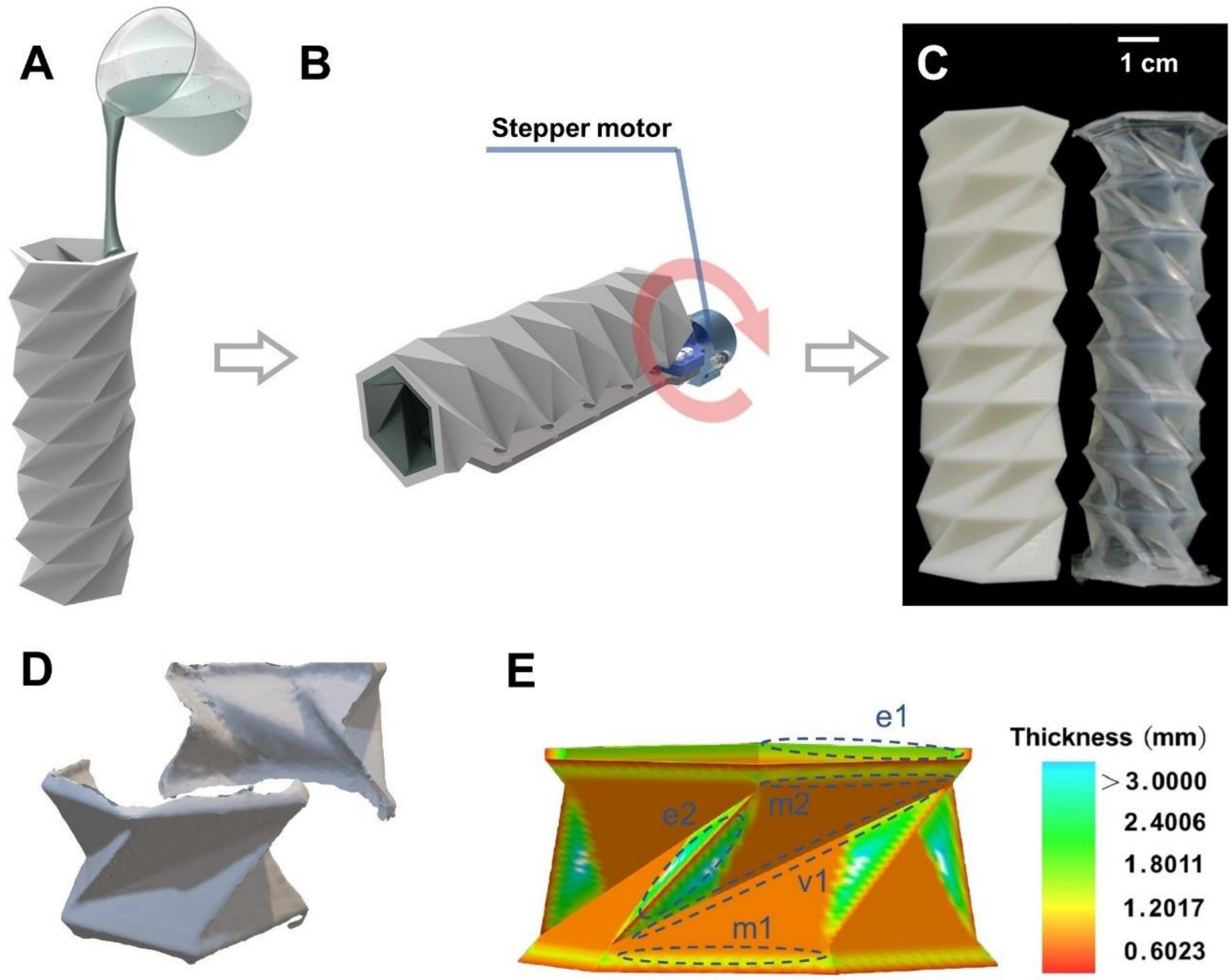

**FIG. 3.** Fabrication process of the OSPAs. (A) Pour the silicone into the mold. (B) Fix the mold on the spin casting machine and rotate it. (C) Demold to obtain the OSPA. (D) Solid silicone model reconstructed using 3D scanning. The model is cut in the middle for observation. (E) Thickness of the OSPA module. e1 and e2 represent the enhanced area of thickness, m1 and m2 refer to the mountain fold, and v1 refers to the valley fold area.

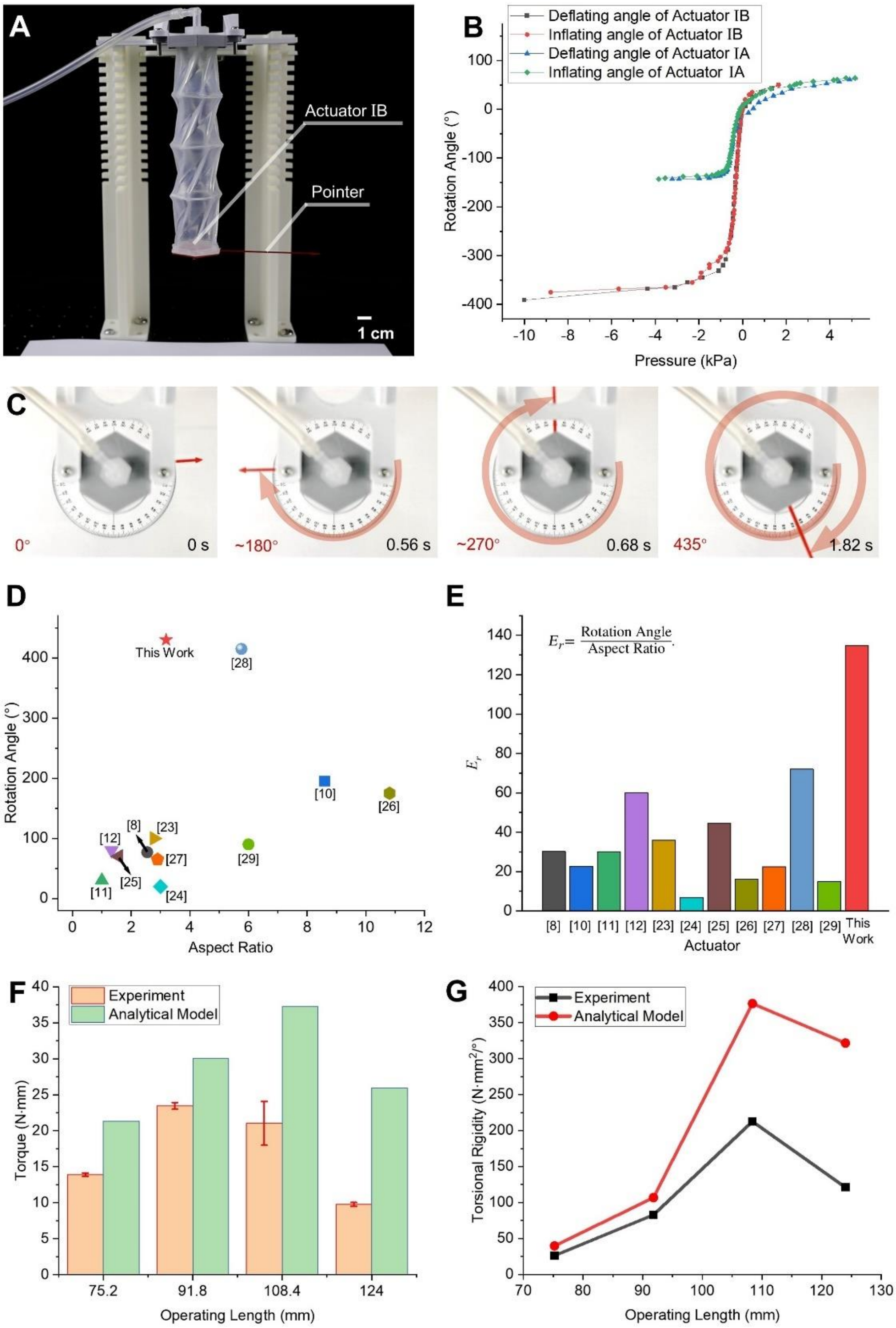

**FIG. 4.** Characterization of the OSPAs. (A) Experimental setup for rotation angle test. (B) The plot of actuator rotation angle versus pressure for the OSPAs. (C) Snapshots of Actuator IB while it rotated by 435°. (D) Comparison of the rotation angle and the amount of change in air pressure between our work and previous fluid-drive soft twisting actuator. (E) Comparison of rotation ratios $E_r$. (F) The results of torque from experiment and analytical model at different operating lengths. (G) The results of rigidity from experiment and analytical model at different operating lengths.

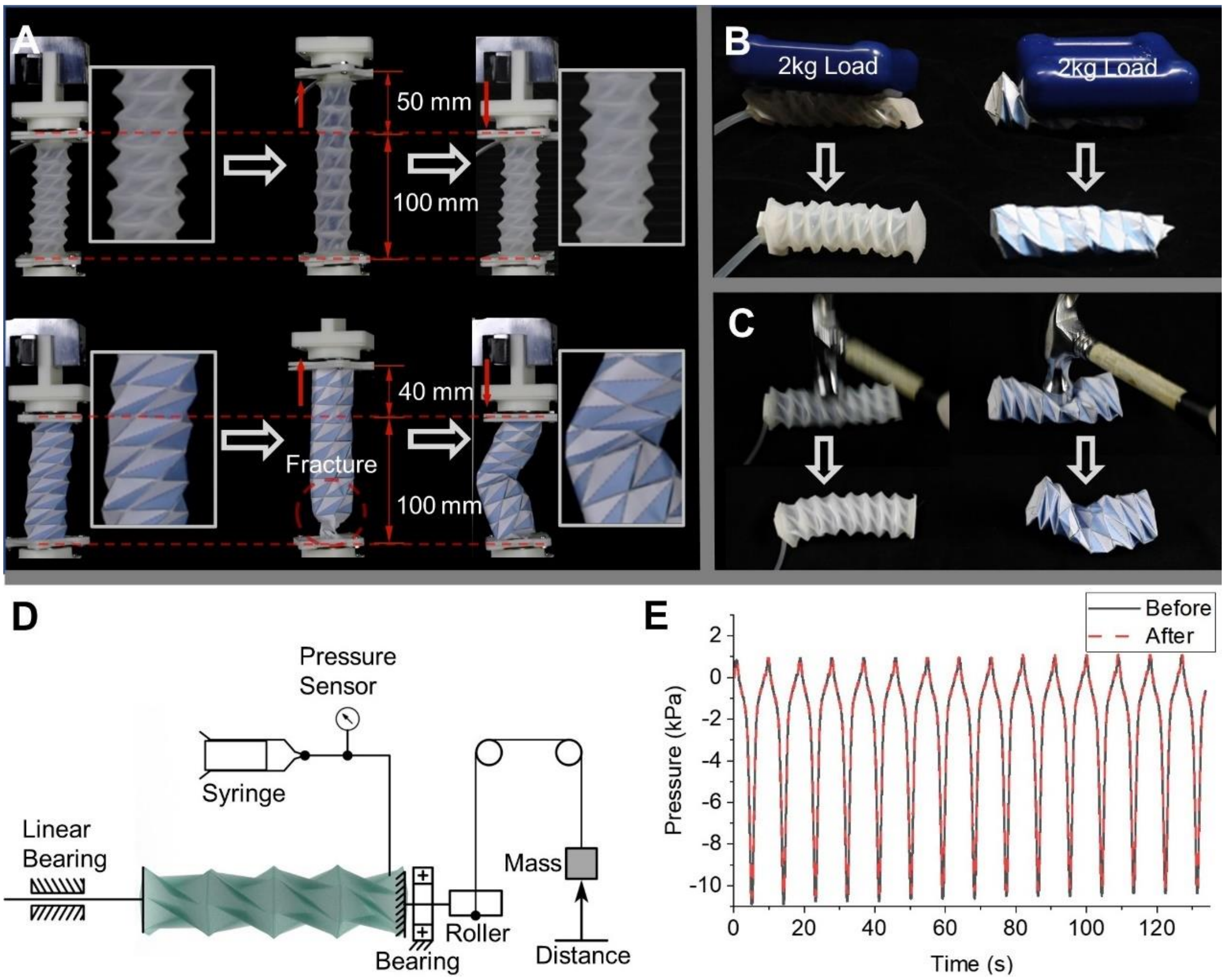

**FIG. 5.** The experiments on the OSPAs. The reliability comparison between OSPAs and paper-made counterparts, including (A) tensile testing, (B) load testing, and (C) impact testing. (D) Schematic of the experimental setup for efficiency and reliability testing. (E) Changes in the internal pressure of the actuator before and after 1000-cycles operation.

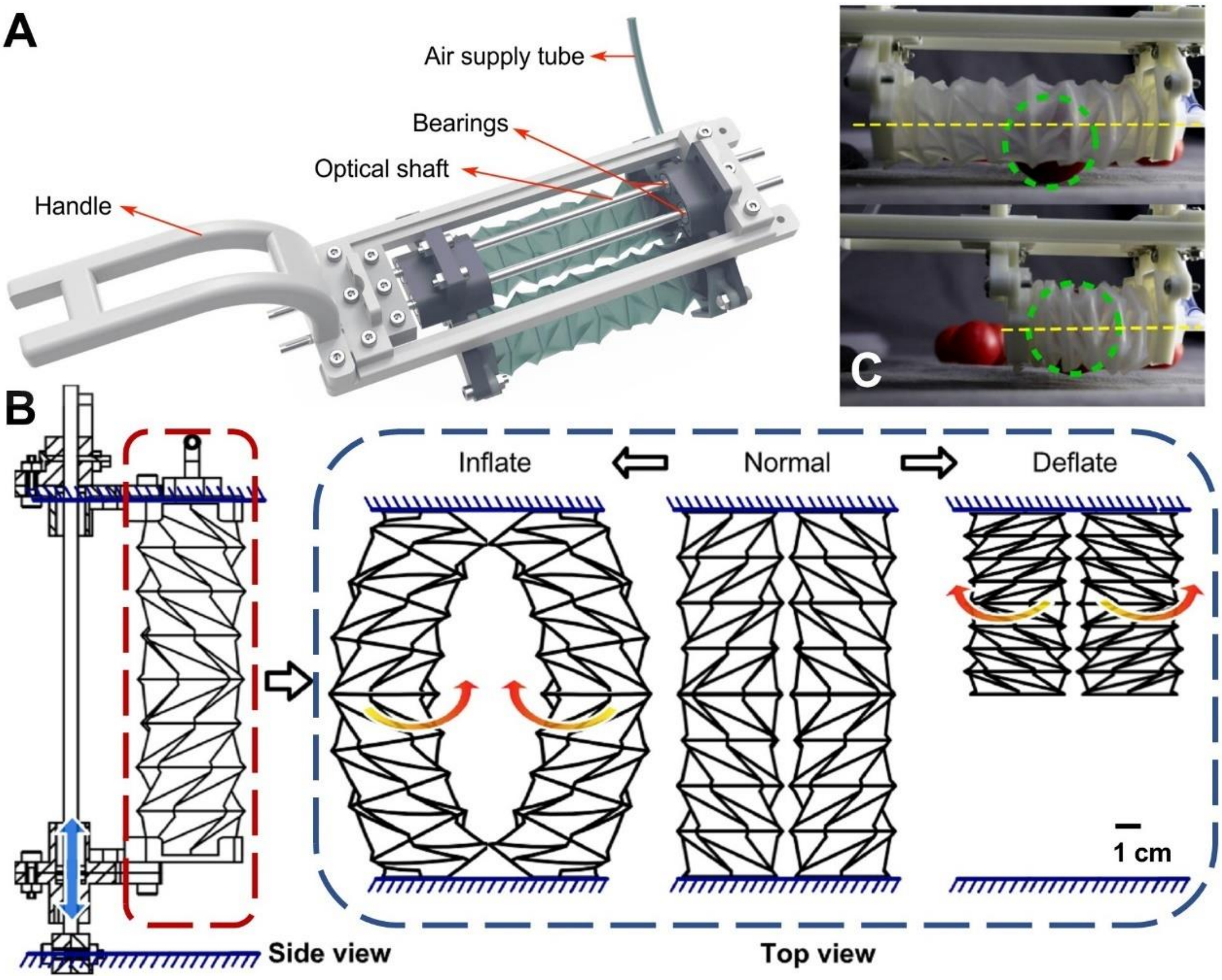

**FIG. 6.** The OSPA gripper. (A) The components of the OSPA gripper. (B) The OSPA gripper working principle. (C) The gripping process of the OSPA gripper, which clamps and lifts the cherry tomato with the rotation motion of the actuators. The green circular dashed frame indicates the position of the cherry tomato and the yellow dashed line represents the centerline of the actuator.

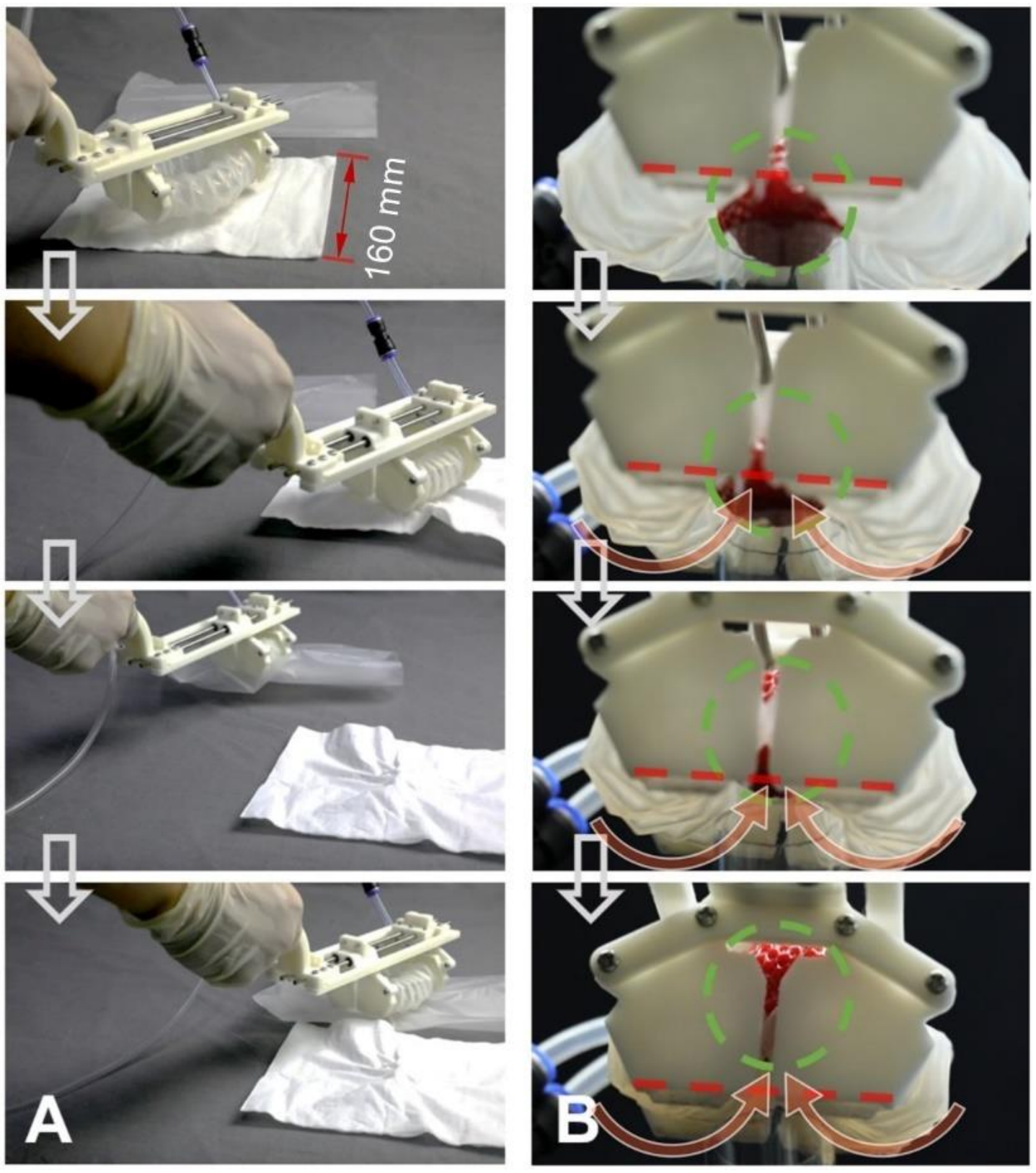

**FIG. 7.** The typical gripping process of the OSPA gripper. (A) Grabbing tissue and plastic bags. (B) Grabbing a strawberry. The bottom view shows the rotation of the actuators.

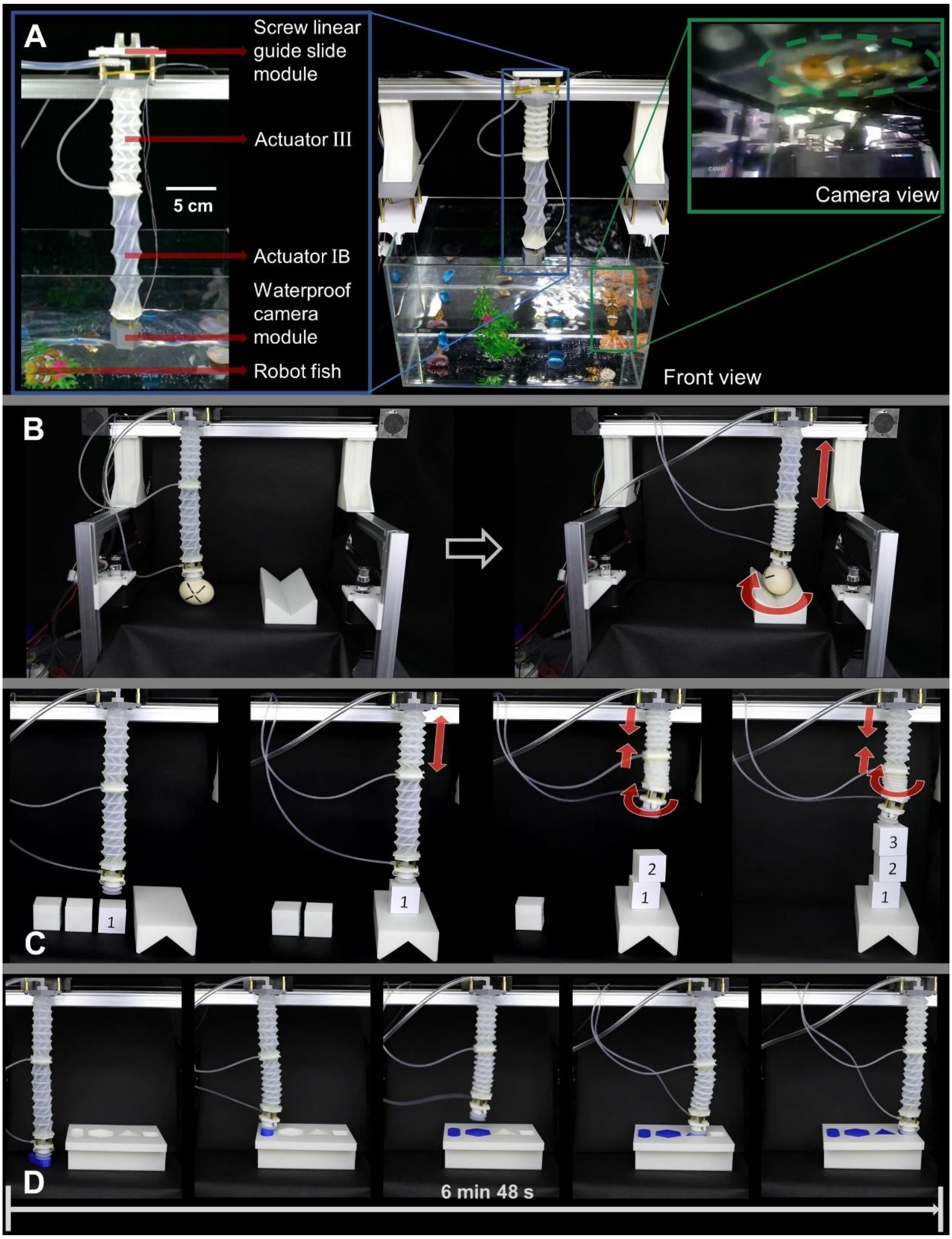

**FIG. 8.** Demonstration on the OPSA arm. (A) The structure of the versatile deployable robot arm. (B) Egg catching demonstration. (C) Tetris game. (D) Picking and placing the shapes in the right slots.

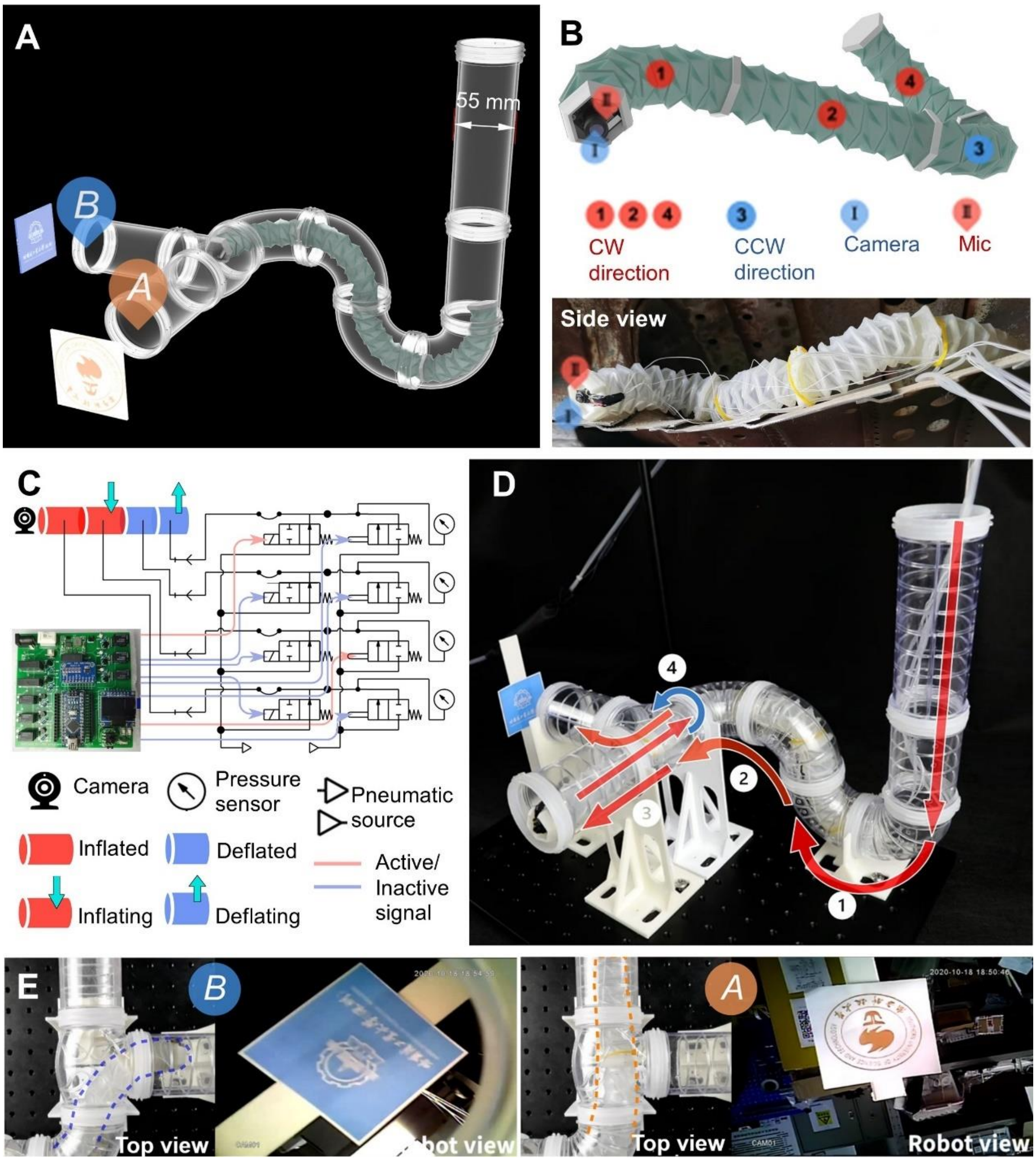

**FIG. 9.** Demonstration of the OSPA snake robot. (A) The structure of the pipe network with two outlets. Each outlet has a university logo. (B) Structural diagram of the soft snake robot, and the side view of the robot when it is located in the gap of a jet engine. (C) The pneumatic circuit. (D) The snapshots of the robot while it is crawling to the exit A in the pipe network and the locomotion path of the robot in the pipe. (E) The snapshots of the robot before and after it rotates the head from one exit to the other one.

## Supplementary Data

Supplementary Data

**Kinematics of an OSPA actuator composed of a series of modules**

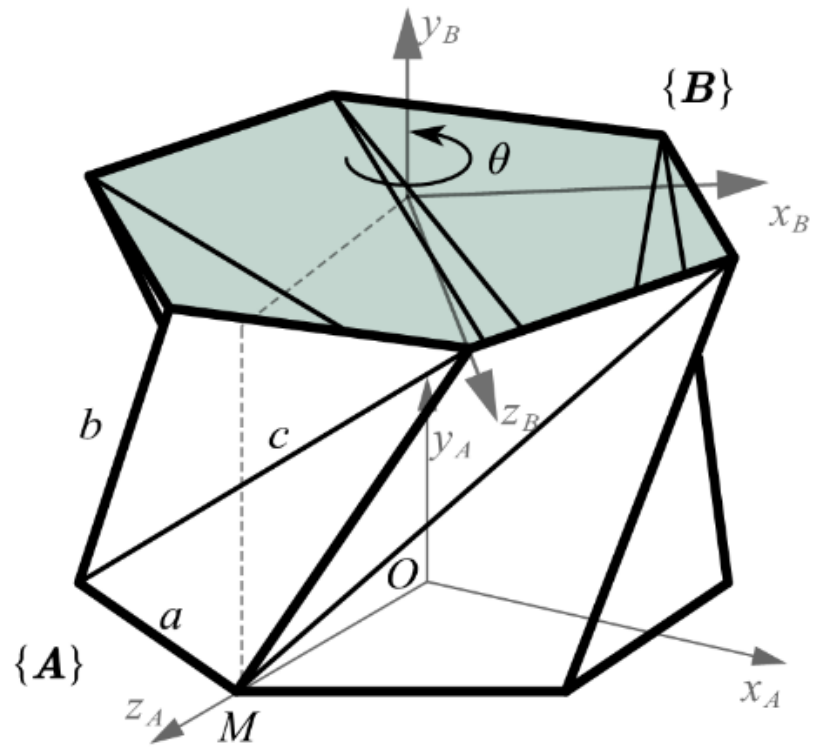

**SUPPLEMENTARY FIG. S1** Coordinate system of the kinematics model of twisting motion.

From elemental rotation matrix of $y$-axis, we have,

$$R_y = \begin{bmatrix} \cos\theta & 0 & \sin\theta \\ 0 & 1 & 0 \\ -\sin\theta & 0 & \cos\theta \end{bmatrix}, \tag{S1}$$

where $\theta$ represents relative rotation angle as shown in Fig. 2(A). There is a homogeneous transformation matrix $H$ which contains relative position and attitude information between the two coordinate systems $\{\boldsymbol{A}\}$ and $\{\boldsymbol{B}\}$,

$$H = \begin{bmatrix} R_y & {}^{A}p_{ab} \\ 0 & 1 \end{bmatrix}, \tag{S2}$$

where ${}^{A}p_{ab}$ is the position vector from $\{\boldsymbol{B}\}$ to $\{\boldsymbol{A}\}$, and ${}^{A}R_B$ refers to the rotational transformation matrix which describes the pose of $\{\boldsymbol{B}\}$ with respect to $\{\boldsymbol{A}\}$. The rigid-body transformation of the top plate with respect to the and bottom plate is given by,

$$H = \begin{bmatrix} \cos\theta & 0 & \sin\theta & 0 \\ 0 & 1 & 0 & b - \dfrac{b}{60^\circ}|\theta| \\ -\sin\theta & 0 & \cos\theta & 0 \\ 0 & 0 & 0 & 1 \end{bmatrix}. \tag{S3}$$

The equation can also be written as,

$$H=\begin{bmatrix} R_y(\theta) & \left(b-\frac{b}{60°}|\theta|\right)\vec{v} \\ \vec{0}^T & 1 \end{bmatrix} \in R^{4\times4}, \tag{S4}$$

where $\vec{v}=[0 \quad 1 \quad 0]^T$.

For the OSPAs with $n$ modules, the transformation matrix from the first OSPA module to the end is,

$${}^0H_n=\begin{bmatrix} \prod_{i=1}^{n} R_y(\theta_i) & \left(nb-\frac{b}{60°}\sum_{i=1}^{n}|\theta_i|\right)\vec{v} \\ \vec{0}^T & 1 \end{bmatrix} \in R^{4\times4}, \tag{S5}$$

where for Actuator IB there is $b=2a$, $h=2\text{a}\cdot\sin\left(\frac{\theta}{2}\right)$. Using these equations, we can analyze the kinematics of an OSPA actuator composed of a series of modules.

### Flexural rigidity and elastic constants

Referring to Silverberg's test method mentioned in the supplementary material[36], several creases 1 mm in length and width were modeled in silicone E615 with two different thicknesses: 1.2 mm and 0.6 mm in FEA simulation. The creases were bent by 60° and the torque was linearly proportional to the bending angular deflection by a constant $k_c$. When the thickness of crease is 1.2 mm, $k_{c1}=2\ N/rad \simeq 3.49\times10^{-2}\ N/°$ and $k_{c2}=0.25\ N/rad \simeq 4.36\times10^{-3}\ N/°$ for thickness of 0.6 mm.

### Volume of the OSPA module and the rotation angle of the folds

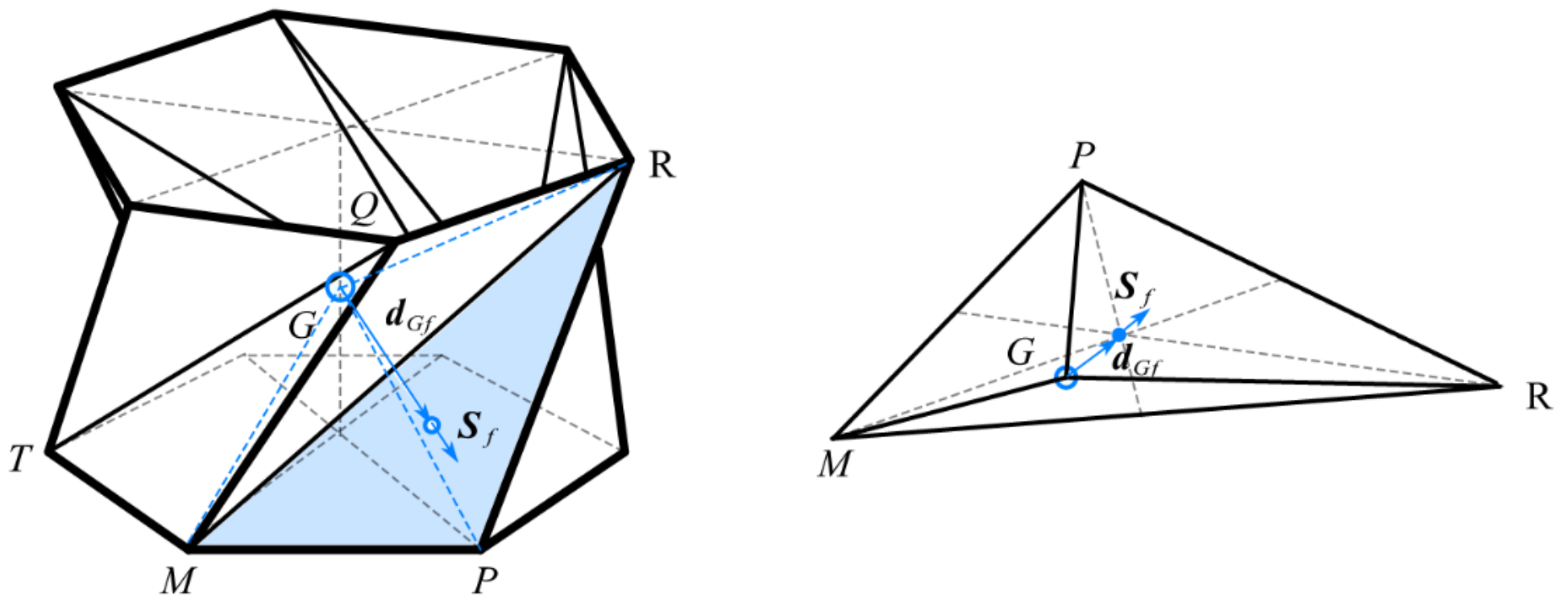

**SUPPLEMENTARY FIG. S2** The sub-element pyramid G-MPR from OSPA module.

Volume of the tetrahedron formed by the geometric center G and the plane MPR can be given as,

$$V_{pyramid}=\frac{1}{3}\boldsymbol{d}_{Gf}\cdot\boldsymbol{S}_f, \tag{S6}$$

where $\boldsymbol{d}_{Gf}$ is the vector from geometric center $G$ to the center of the polygon $f$ and $\boldsymbol{S}_f$ is the normal vector of plane MPR.

In Fig. 2(A), the coordinates of the four points MQPR and geometric center $G$ on truss structure can be written as,

$$\begin{cases} M(0,0,a) \\ P\left(\dfrac{\sqrt{3}}{2}a, 0, \dfrac{a}{2}\right) \\ Q(a \cdot \sin\theta_u, h, a \cdot \cos\theta_u) \\ R(a \cdot \sin(\theta_u + 60°), h, a \cdot \cos(\theta_u + 60°)) \\ G\left(0, \dfrac{h}{2}, 0\right) \end{cases} . \tag{S7}$$

Since OPSA consists of soft materials, we may assume that the closure condition applies to most parameters. In this way, the relationship between the volume of the OSPA module and $\theta_u$ can be given as,

$$\begin{aligned} V_C(\theta_u) &= 12 \cdot V_{GRMP} + 2 \cdot \left(\frac{1}{3} \cdot \frac{h}{2} \cdot S_{bottom}\right) \\ &= 4 \cdot (\mathrm{proj}_{\boldsymbol{s}_f} \boldsymbol{GP}) \cdot S_{MPR} + \frac{h}{3} \cdot S_{bottom} \\ &= \frac{\sqrt{2}\, a^2\, b\left(2\sigma_1 + \sqrt{3}\right)\sigma_2 \sqrt{4\,|\sigma_3|\;|a\, b\, \sigma_1|^2 + 4\,|\sigma_3|\;|a\, b\, \sin(\theta_u + 30^\circ)|^2 + |a|^4\;|b|^2\;\left|2\sigma_1 - \sqrt{3}\right|^2}}{4\,|b|\sqrt{a^4\cos(\theta_u)^2 - 2a^4 + 2a^2b^2 + a^4\cos(\theta_u) + \sqrt{3}\,a^4\sin(\theta_u) - \sqrt{3}\,a^4\cos(\theta_u)\sin(\theta_u)}} \\ &\quad + \frac{\sqrt{3}\, a^2\, \bar{\sigma}_2\, \bar{b}}{2} \end{aligned} , \tag{S8}$$

where $\sigma_1 = \cos(\theta_u + 30^\circ)$, $\sigma_2 = \sqrt{\dfrac{\sigma_3}{b^2}}$, $\sigma_3 = b^2 - 2a^2 + 2a^2\cos(\theta_u)$ and $S_{bottom}$ represent the area of the hexagon base.

The rotation angles of the folds QR, QM, RM can be written as,

$$\theta_{QM} = \arccos\left(\frac{|(\boldsymbol{QM} \times \boldsymbol{QR}) \cdot (\boldsymbol{QM} \times \boldsymbol{MT})|}{|\boldsymbol{QM} \times \boldsymbol{QR}| \cdot |\boldsymbol{QM} \times \boldsymbol{MT}|}\right), \tag{S9}$$

$$\theta_{QR} = \arccos\left(\frac{|(\boldsymbol{QM} \times \boldsymbol{QR}) \cdot (\boldsymbol{n}_s)|}{|\boldsymbol{QM} \times \boldsymbol{QR}| \cdot |\boldsymbol{n}_s|}\right), \tag{S10}$$

$$\theta_{RM} = \arccos\left(\frac{|(\boldsymbol{QM} \times \boldsymbol{QR}) \cdot (\boldsymbol{RM} \times \boldsymbol{MP})|}{|\boldsymbol{QM} \times \boldsymbol{QR}| \cdot |\boldsymbol{RM} \times \boldsymbol{MP}|}\right), \tag{S11}$$

where $\boldsymbol{n}_s = [0, 1, 0]$.

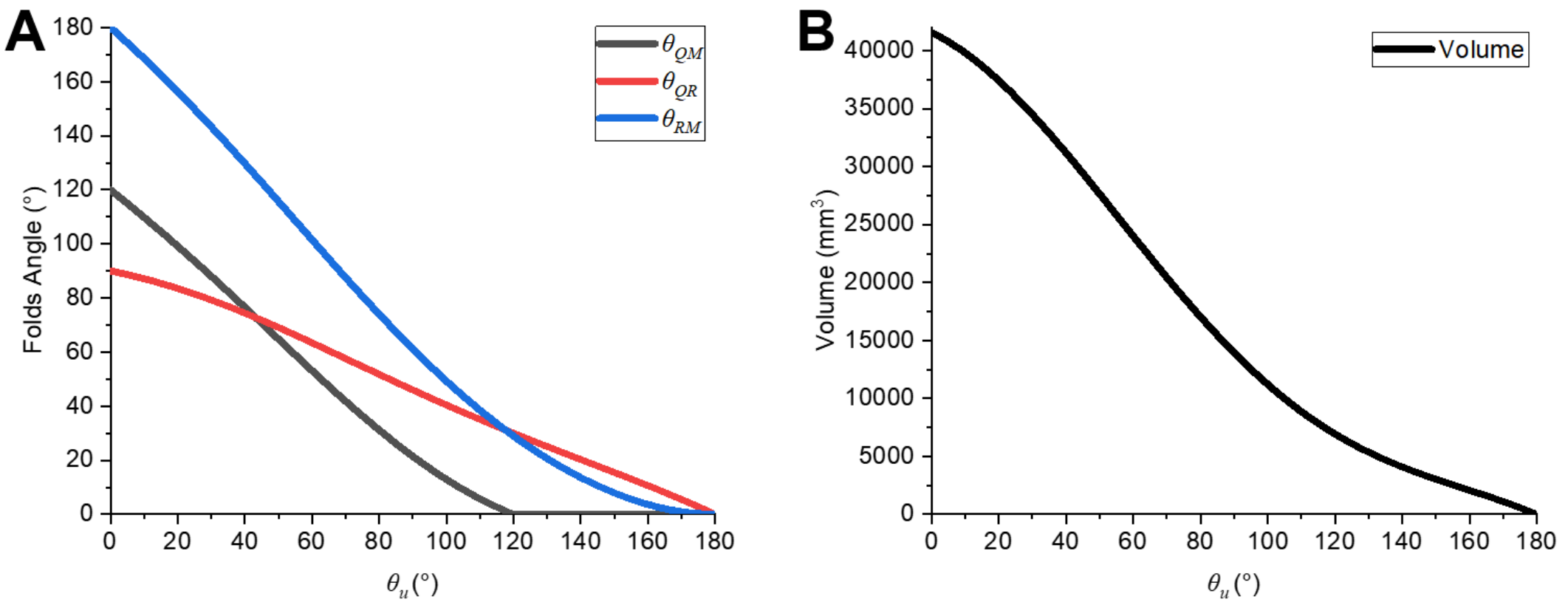

**SUPPLEMENTARY FIG. S3** (A) The rotation angles of the folds QM, QR, RM are plotted as a function of relative rotation angle $\theta_u$. (B) Volume of the OSPA module Type IB ($V_C$) is plotted as a function of relative rotation angle $\theta_u$.

**SUPPLEMENTARY Table S1** Fitting results

| Material | Parameters |
|---|---|
| Ecoflex 00-30 | $C_{10} = 0.00364188$<br>$C_{20} = 0.000573251$<br>$C_{30} = -3.93058e-06$ |
| E615 | $C_{10} = 0.0727207$<br>$C_{20} = 0.00527073$<br>$C_{30} = -7.73102-05$ |
| Mixture (Dragonskin 30: Ecoflex 00-30=1:1) | $C_{10} = 0.0683405$<br>$C_{20} = 0.00958809$<br>$C_{30} = -0.000363852$ |

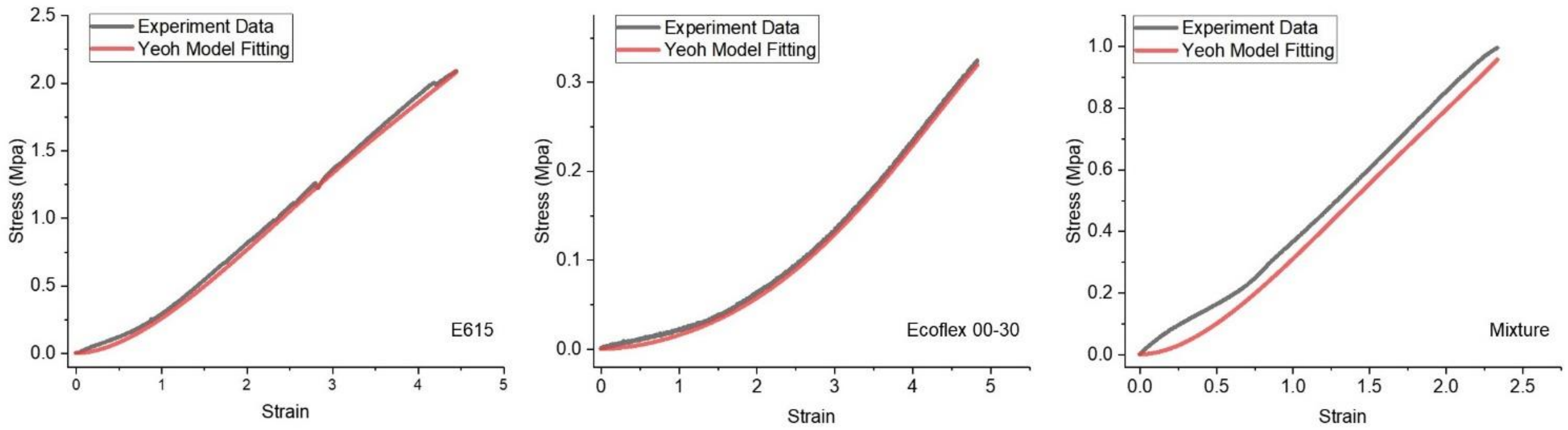

**SUPPLEMENTARY FIG. S4** Stress-strain curves of three materials. (A) E615 From Hongyejie Inc. (B) Ecoflex 00-30 from Smooth-on. (C) The mixed components of Dragonskin 00-30 and Ecoflex 00-30 with a 1:1 weight ratio. Uniaxial tensile tests were performed more than five times for each material, using the tensile testing machine from MTS (Model 42). The standard throughout the whole experiment was referred to ASTM D412 rubber tensile test which is the most common standard for determining the tensile properties of (silicone) rubber. Dimensions of testing specimens are as defined in Type C of D412.

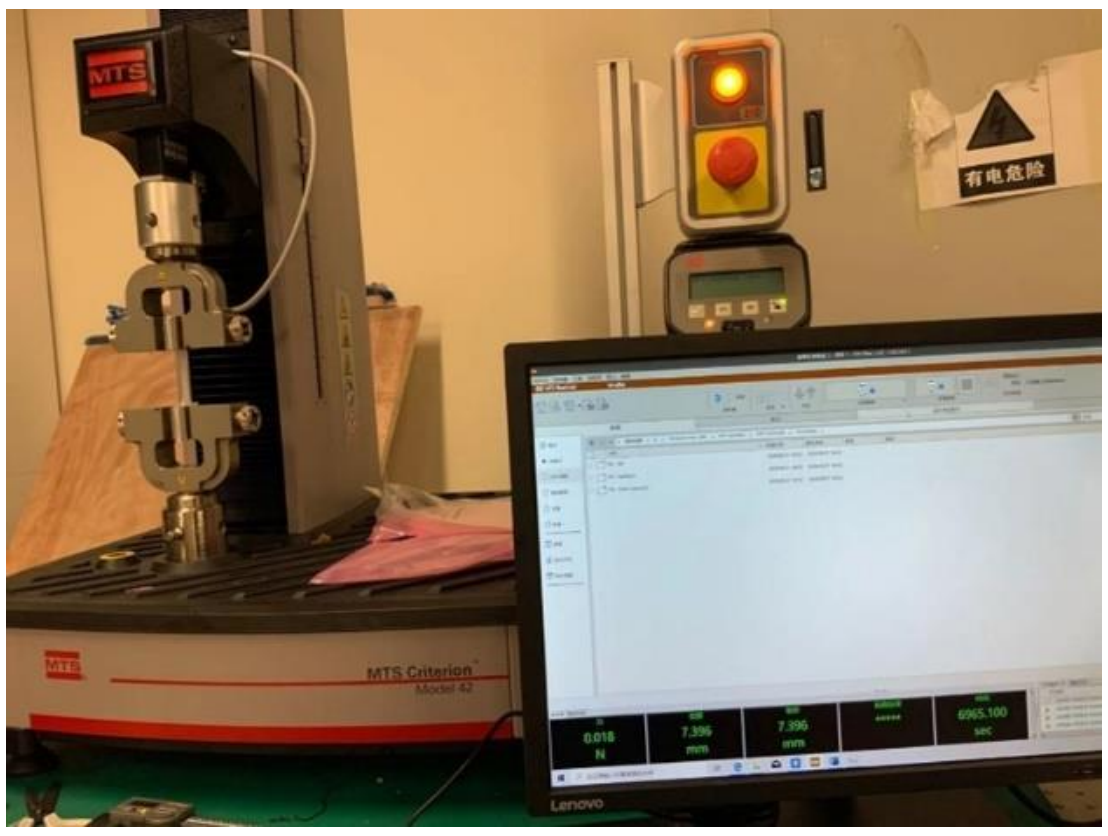

**SUPPLEMENTARY FIG. S5**  Experimental setup for measuring stress-strain curves.

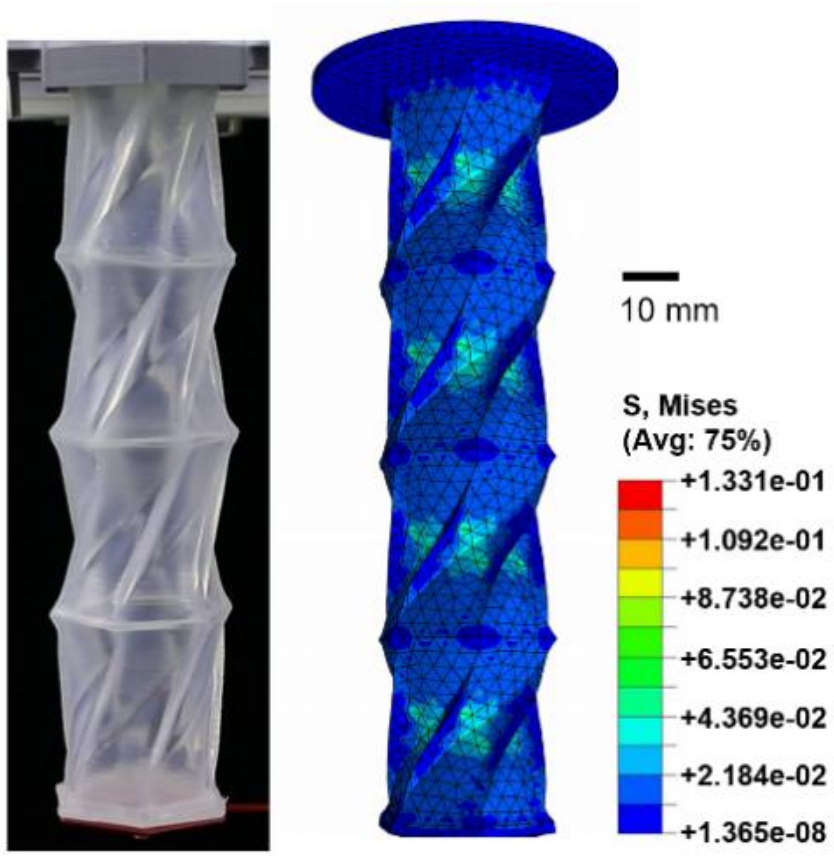

**SUPPLEMENTARY FIG. S6**  Actuator IB of E615 at a positive pressure of about 2 kPa in the experiment and the FEA simulation.

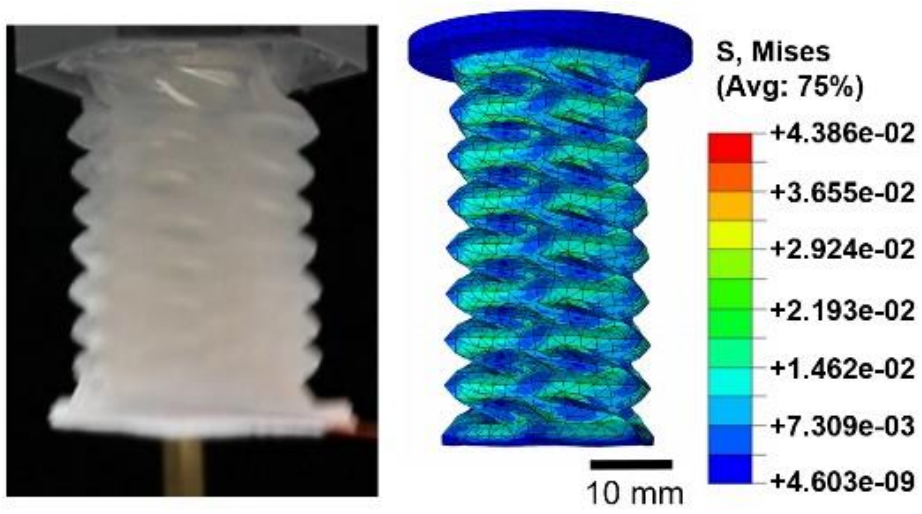

**SUPPLEMENTARY FIG. S7**  Actuator IA of E615 at a negative pressure about -3 kPa in the experiment and the FEA simulation.

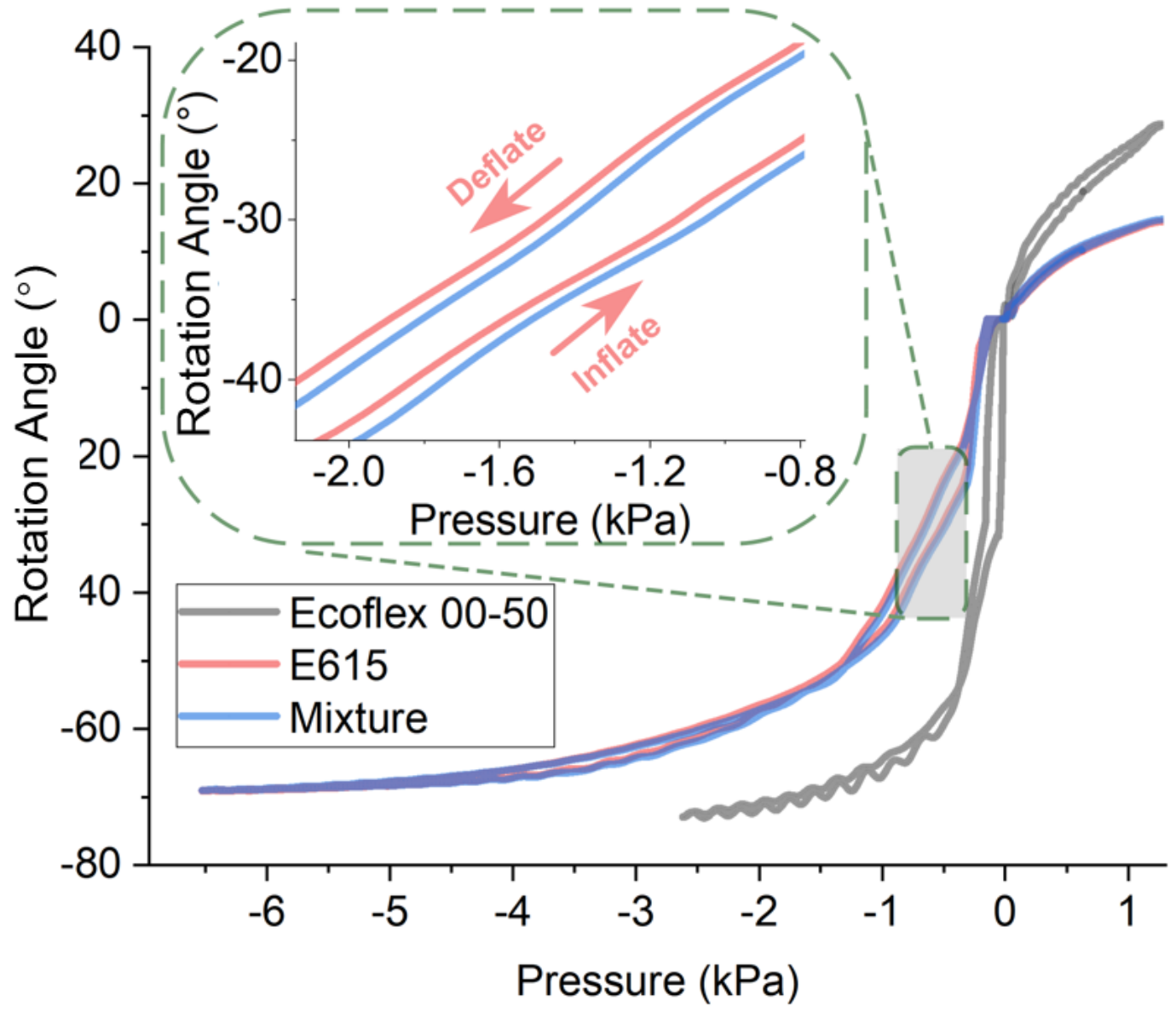

**SUPPLEMENTARY FIG. S8** Simulation results of the variation of rotation angle with pressure for the module Type IB made of three different silicone.

**SUPPLEMENTARY TABLE S2.** Coating process of the OSPAs

| Step | Tasks | Time | Note |
|---|---|---|---|
| 1 | Prepare the silicone precursor, weight ratio A:B=1:1 and stir with a coffee stick for 2-3 mins | 5-8 mins | This step does not require vacuum defoaming |
| 2 | Evenly dip-coat the inside of the mold in sequence, fix the finished dip-coated mold to the inner frame of the spin-coater and turn on the spin-coating machine | 20 mins | Dip coating for approx. 5 mins, and spin coating need to be at room temperature |
| 3 | Remove the mold from the spin coater and transfer it to the vacuum oven, close the balance valve and turn on the vacuum pump to defoam | ~ 5 mins | |
| 4 | Turn off the vacuum pump and slowly unscrew the balance valve | 5 mins | |
| 5 | After the previous step, the mold should be immediately transferred to a flat surface (e.g. a square Petri dish) and cured by heating (45°C) for 30 mins | 30 mins | |
| 6 | Repeat steps 2-5, depending on the desired thickness, usually 3 times | — | The direction of the mold needs to be changed each time after it is dipped. |
| 7 | Evenly coat a piece of non-woven fabric slightly smaller than a square Petri dish with silica gel, place the mold on top of the silica gel, seal its | — | |

| | | |
|---|---|---|
| | side, vacuum defoam and heat, and repeat steps 3-5 | |
| 8 | Using the twist-off demolding method, peel the cured silicone from the mold, and insert the internal support skeleton from the other side (if it is needed) | — |
| 9 | Referring to step 7, seal the other side and heat to cure | 30 mins |
| 10 | Takeout the cured product and check air tightness | — |
| 11 | Cut a small hole in the center of one side and use the connector to seal with silicone | 30 mins-1 hour |
| 12 | Waiting for curing, finished | — |

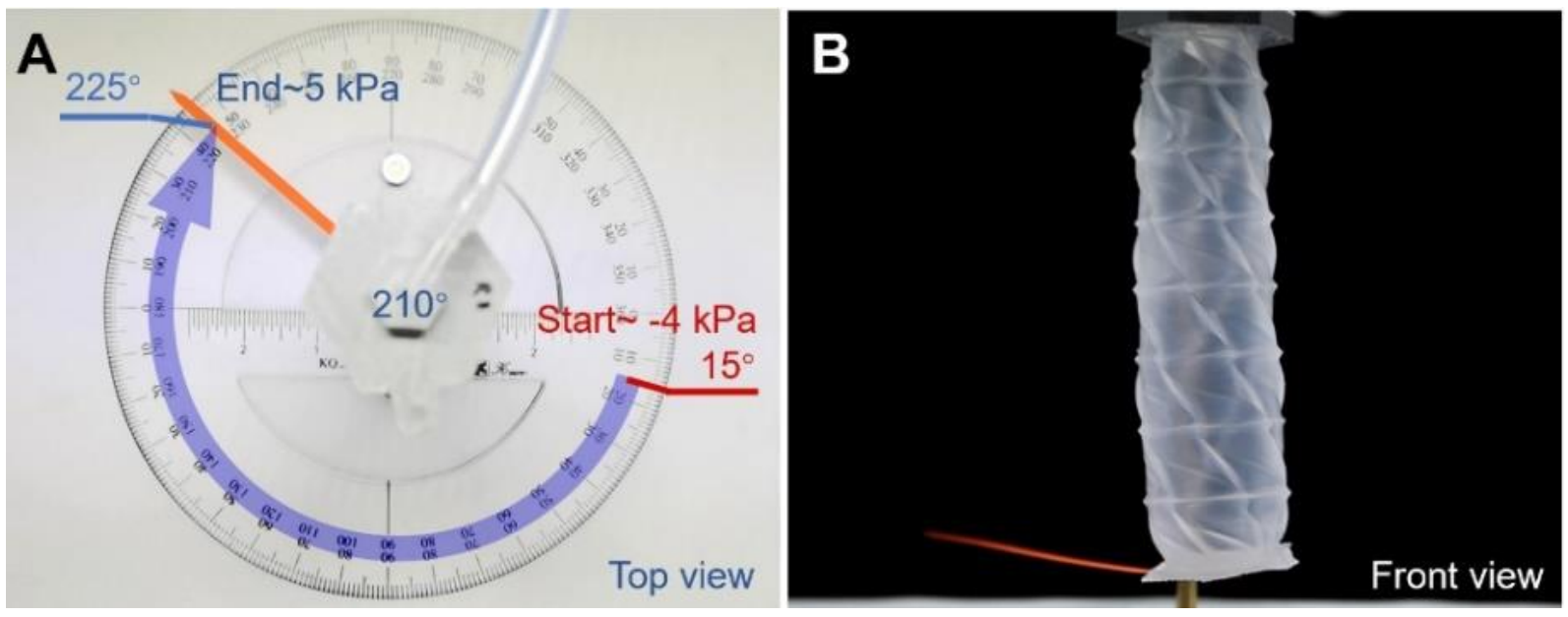

**SUPPLEMENTARY FIG. S9** Experimental setup for measuring rotation angle (Actuator IA).

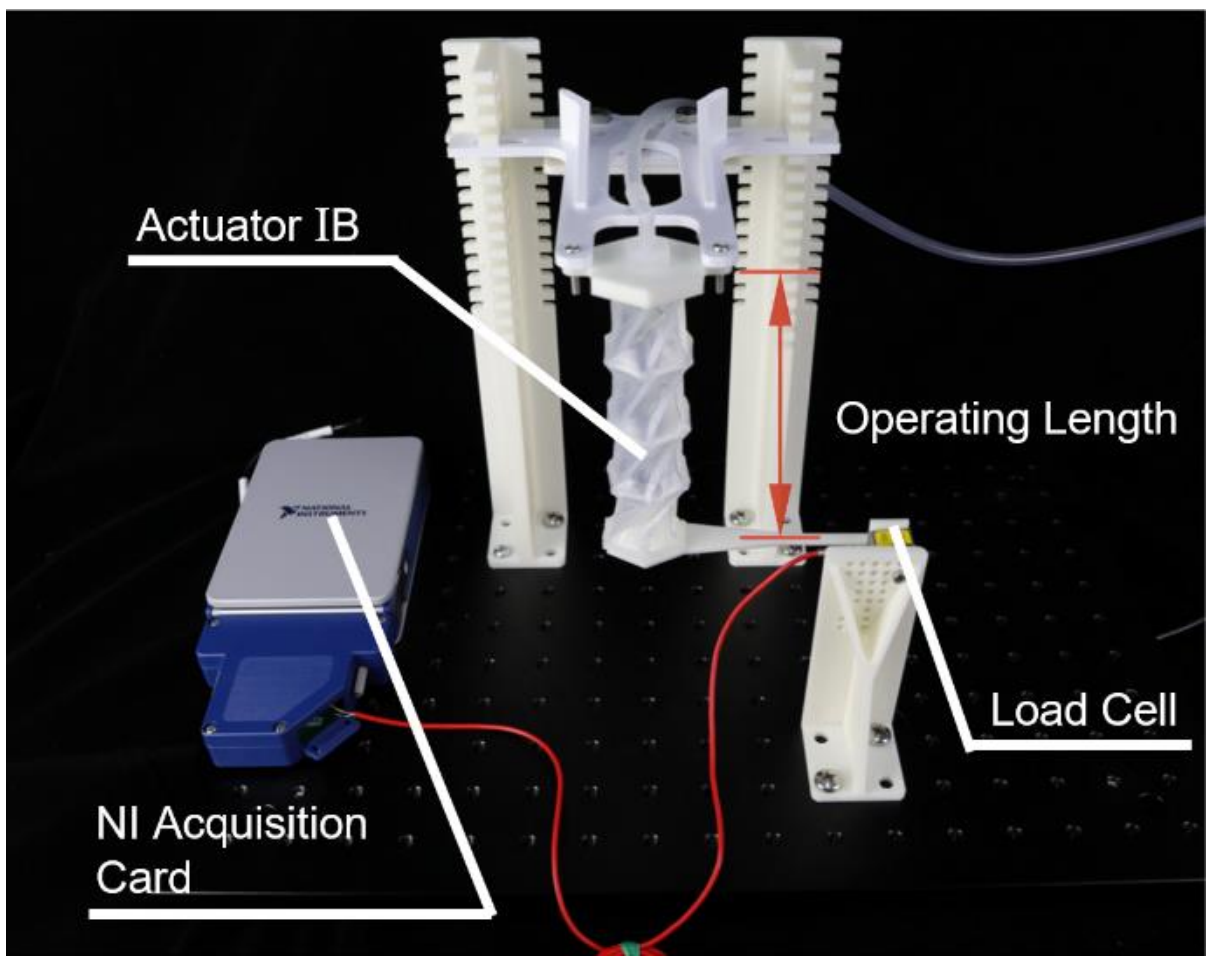

**SUPPLEMENTARY FIG. S10** Experimental setup for torque testing (Actuator IB).

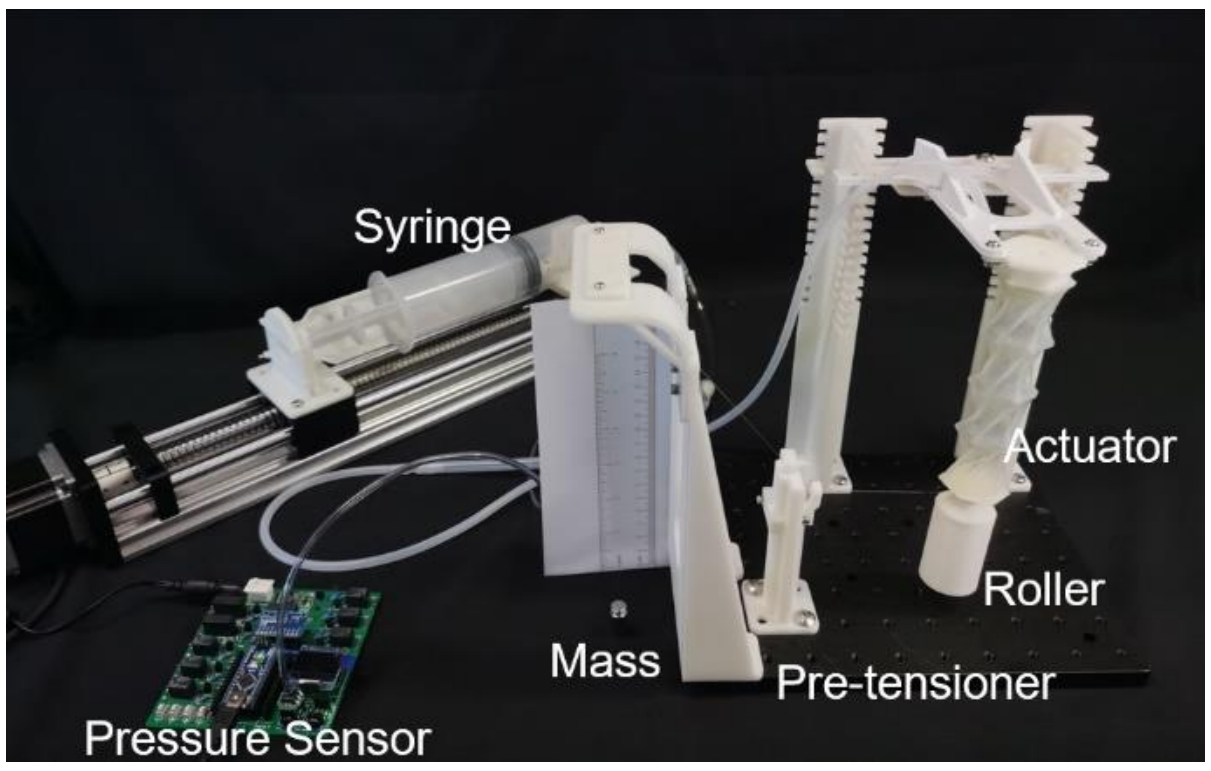

**SUPPLEMENTARY FIG. S11** Experimental setup for life testing (Actuator IB).

**OSPA's internal support skeleton**

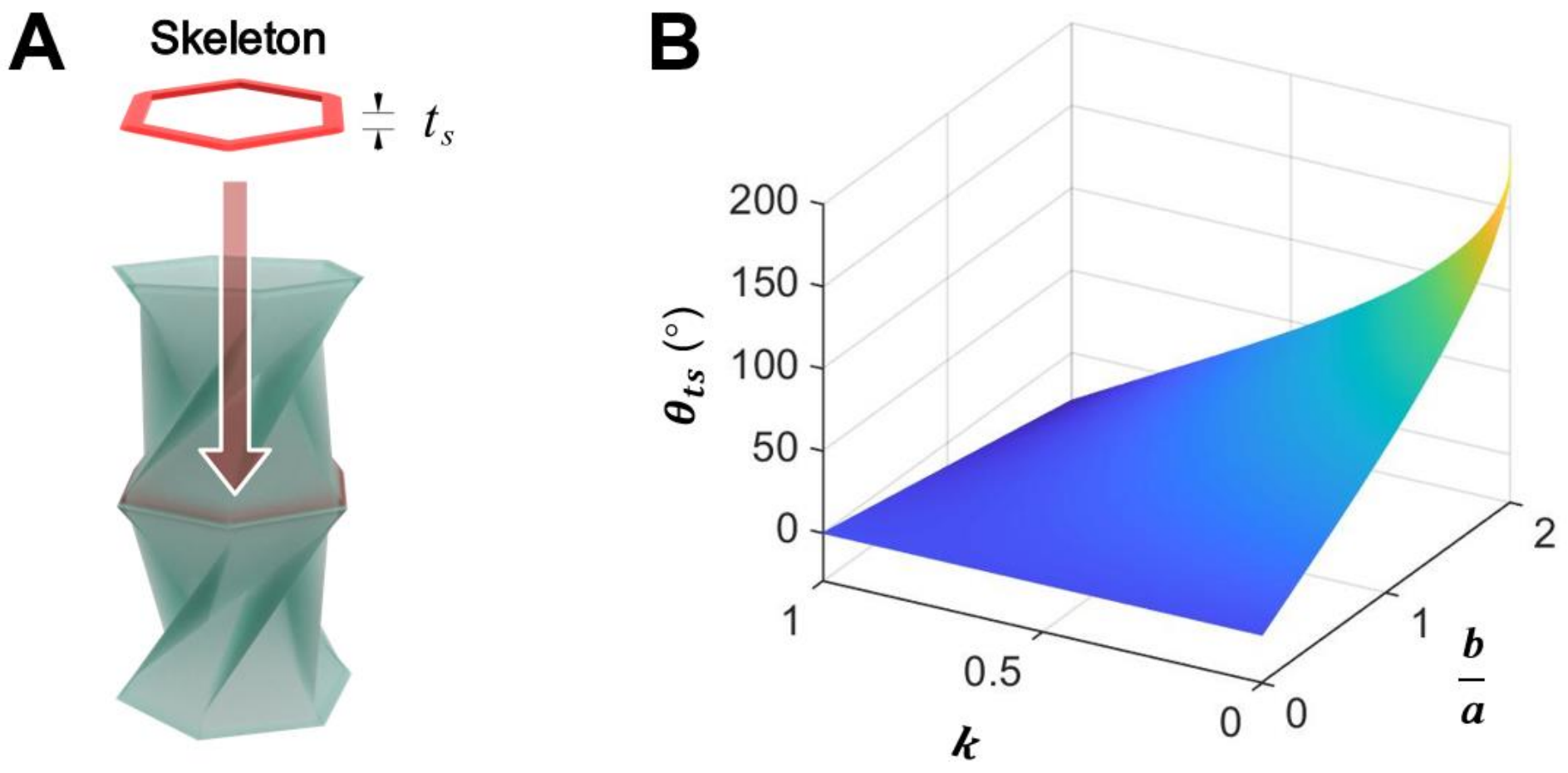

**SUPPLEMENTARY FIG. S12** (A) OSPA's internal support skeleton. (B) The influence of $k$ and $b/a$ on the maximum rotation angle when the skeleton is inserted ($\theta_{ts}$).

Occasionally, under extreme loads, the actuator may unfavorably collapse in the radial direction. To avoid this deformation and generate a larger load capacity, we insert internal skeletons made by Polylactic Acid (PLA) or Thermoplastic polyurethanes (TPU) to support the structure.

When the support structure is inserted, the thickness of the skeleton $t_s$ influences the movement of the OSPA module. Assuming that the surface of the skeleton and the OSPA do not slide relative to each other during the actuation, we can obtain the following equation,

$$\theta_{ts} = 2\arcsin\left(\frac{b}{2a}\left(1 - \frac{k}{\sin\delta_0}\right)\right), \tag{S12}$$

where $t_s$ is the thickness of the skeleton and $k = \frac{t_s}{b}$. It can be seen that keeping the thickness of the support structure thin has only a modest effect on the maximum rotation angle. The thickness of the internal support skeleton we used is usually less than $b/20$, which minimizes the effect on the overall kinematics.

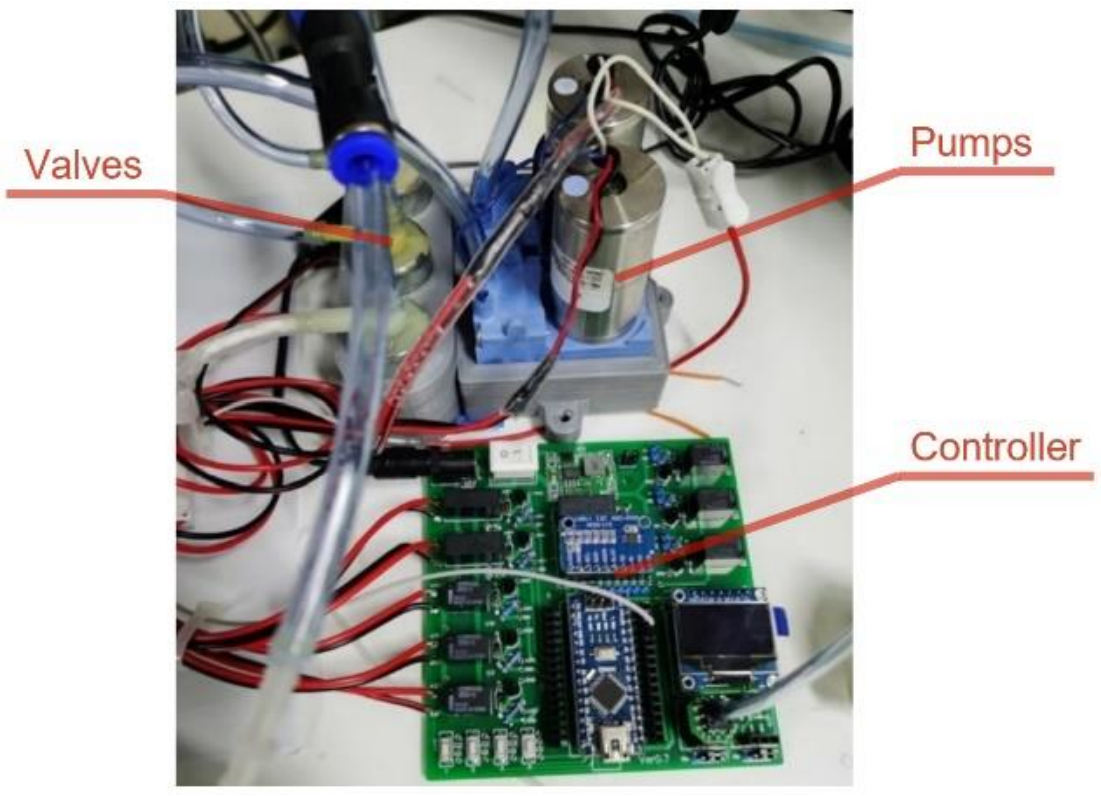

**SUPPLEMENTARY FIG. S13** A pneumatic controller with two pumps and four valves. Self-designed pneumatic controller based on Arduino nano, up to three independent actuators can be driven simultaneously, with 16bit ADC sampling, air pressure monitoring and other functions.